\documentclass{article}

\usepackage{microtype}
\usepackage{graphicx}
\usepackage{subfigure}
\usepackage{booktabs} % for professional tables
\usepackage{times}
\usepackage{epsfig}
\usepackage{amsmath}
\usepackage{amssymb,amsthm}
\usepackage{bm}
\usepackage{amsfonts}
\usepackage{mathrsfs}
\usepackage[mathscr]{eucal}
 \newcommand{\M}[1]{{\bm{\mathbf{\MakeUppercase{#1}}}}} % matrix
\newcommand{\vx}{\mathbf{x}}
\newcommand{\vy}{\mathbf{y}}

\newcommand{\mU}{\mathbf{U}}
\newcommand{\mX}{\mathbf{X}}
\newcommand{\mY}{\mathbf{Y}}

\newcommand{\bmX}{\mathbf{\overline{X}}}

\newcommand{\vz}{\mathbf{z}}

\newcommand{\cS}{\mathcal{S}}
\newcommand{\cO}{\mathcal{O}}

\newcommand{\vidloss}{\mathcal{L}_C}
\newcommand{\dataloss}{\mathcal{L}_{\mathcal{D}}}
\newcommand{\reploss}{\mathcal{L}_{R}}
\newcommand{\lot}{\mathcal{L}_{OT}}
\newcommand{\lgan}{\mathcal{L}_{G}}
\newcommand{\ladv}{\mathcal{L}_{A}}

\newcommand{\grassmann}{\mathcal{G}}

\newcommand{\kernel}{\mathbf{K}}

\newcommand{\labels}{\mathcal{L}}
\newcommand{\dataset}{\mathcal{D}}
\newcommand{\classifier}{C}
\newcommand{\fcl}{\zeta}
\newcommand{\eye}{I}

\newcommand{\hinge}[1]{\left[{#1}\right]_+}
\newcommand{\fnorm}[1]{\left\|{#1}\right\|_F}

\newcommand{\prob}[1]{P_{#1}}
\newcommand{\pD}{\prob{\dataset}}
\newcommand{\pY}{\prob{\mY}}

\newcommand{\fU}{f_{\mU}}
\newcommand{\nuY}{\nu_{\mY}}
\newcommand{\gth}{g_{\theta}}

\newcommand{\expect}[1]{\mathbb{E}_{#1}}

\newcommand{\reals}[1]{\mathbb{R}^{#1}}
\DeclareMathOperator{\dist}{dist}

\newcommand{\enorm}[1]{\left\|{#1}\right\|}
\newcommand{\inner}[1]{\left\langle{#1}\right\rangle}
\DeclareMathOperator{\argmin}{arg\,min}
\newcommand{\set}[1]{\left\{#1\right\}}
\newcommand{\seq}[1]{\left\langle{#1}\right\rangle}

\newcommand{\sqbrack}[1]{\left[#1\right]}

\newcommand{\normal}{\mathcal{N}}

\DeclareMathOperator{\FCN}{FCN}
\DeclareMathOperator{\ReLU}{ReLU}
\DeclareMathOperator{\softmin}{softmin}

\newcommand{\V}[1]{{\bm{\mathbf{\MakeLowercase{#1}}}}} % vector
\newcommand{\mb}[1]{\mathbb{#1}}
\newcommand{\mc}[1]{\mathcal{#1}}
\newtheorem{thm}{Theorem}
\usepackage[mathscr]{eucal}
\usepackage{bm}
\usepackage{multirow}

\usepackage{hyperref}

\usepackage[accepted]{icml2020}

\icmltitlerunning{Adversarially-Contrastive Optimal Transport}

\begin{document}

\twocolumn[
\icmltitle{Representation Learning via Adversarially-Contrastive Optimal Transport}




\begin{icmlauthorlist}
\icmlauthor{Anoop Cherian}{MERL}
\icmlauthor{Shuchin Aeron}{TUFTS}
\end{icmlauthorlist}

\icmlaffiliation{MERL}{Mitsubishi Electric Research Labs, Cambridge, MA.}
\icmlaffiliation{TUFTS}{Tufts University, Medford, MA}

\icmlcorrespondingauthor{Anoop Cherian}{cherian@merl.com}

\icmlkeywords{contrastive learning, optimal transport, adversarial learning, representation learning, video classification}

\vskip 0.3in
]

\printAffiliationsAndNotice{}  % leave blank if no need to mention equal contribution

\begin{abstract}
    In this paper, we study the problem of learning compact (low-dimensional) representations for sequential data that captures its implicit spatio-temporal cues. To maximize extraction of such informative cues from the data, we set the problem within the context of contrastive representation learning and to that end propose a novel objective via optimal transport. Specifically, our formulation seeks a low-dimensional subspace representation of the data that jointly (i) maximizes the distance of the data (embedded in this subspace) from an adversarial data distribution under the optimal transport, a.k.a. the Wasserstein distance, (ii) captures the temporal order, and (iii) minimizes the data distortion. To generate the adversarial distribution, we propose a novel framework connecting Wasserstein GANs with a classifier, allowing a principled mechanism for producing good negative distributions for contrastive learning, which is currently a challenging problem. Our full objective is cast as a subspace learning problem on the Grassmann manifold and solved via Riemannian optimization. To empirically study our formulation, we provide experiments on the task of human action recognition in video sequences. Our results demonstrate competitive performance against challenging baselines.
\end{abstract}

\section{Introduction}
Recent advancements in deep neural network architectures \cite{greff2016lstm,merity2017regularizing,sutskever2014sequence,zilly2017recurrent,varol2017long,RNN1,NIPS2019_9332} have resulted in significant progress towards our ability to model and reason over sequential data. However, this problem is far from considered solved and continues to be challenging, especially in high-dimensional spatio-temporal settings. There are several practical issues that lead to this difficulty, notably (i) most of the highly successful neural network models operate on data of a fixed length (such as images), however temporally-evolving data, such as for example the frame-level features from a video recognition task, could be of arbitrary temporal length, and (ii) the data may be entangled with nuisance factors, such as for example, features corresponding to temporally-evolving background clutter, which may make inference difficult. While, it may be possible to extend popular deep architectures to address these challenges, they may be computationally heavy or require large-scale annotated data, which may be difficult to gather. We refer the reader to several papers \cite{Dollar,VarolLS16,WangG15a,Tran2017, Wu2017, Girdhar_ICCV19} illustrating the wide range of approaches undertaken to tackle this challenging problem. 

In this paper, we address these issues by learning a compact representation of a given video sequence of arbitrary length, that maximally captures the spatio-temporal information, while at the same time can be effectively fed to a light weight classifier for action recognition. We approach this representation learning problem from the perspective of contrastive learning \cite{SaunshiPAKK19, Oord2018, Avishek_ACE, Tschannen2019, Wang_Cherian2018} that has recently emerged as a flexible yet powerful tool for learning generic representations for high dimensional data, using positive and negative examples. The key idea in contrastive learning is to produce a representation that is closer to the positive (given data) examples, and farther from the negatives. Usually, the negative examples are randomly picked. However, it is usually seen that the performance of these algorithms heavily depend on the choice of the negatives and their \emph{pairings} with the positives. ~\cite{arora2019theoretical,Avishek_ACE}.

\begin{figure*}[h]
    \centering
    \includegraphics[width=16cm,trim={1cm 4cm 0.7cm 1cm},clip]{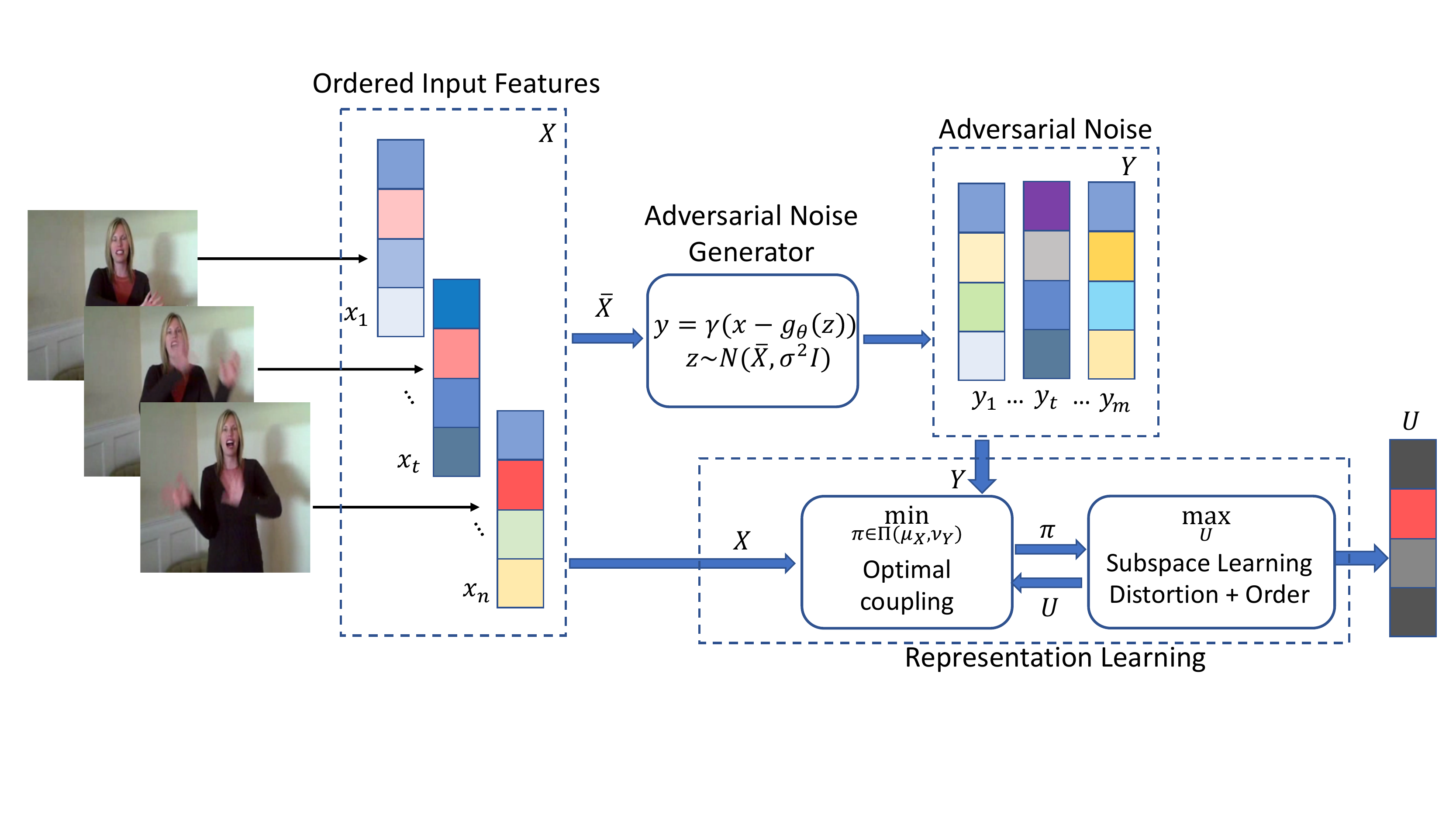}
    \caption{Our overall architecture. The input frames are first encoded into a set $\mX$ of ordered feature vectors $\vx_t$ using a (pre-trained) neural network. These features are then used in an adversarial noise generator (implemented using a Wasserstein GAN trained for adversarial losses) to generate a set $\mY$ of adversarial noise samples. Next, we use $\mX$ and $\mY$ in a joint optimal transport and representation learning formulation that tries to (i) minimize the optimal coupling $\pi$ between the two sets, while (ii) also learn a subspace $\mU$ that maximizes the distance between projections of $\mX$ onto $\mU$ and the adversarial noise $\mY$. This latter cost also includes distortion and ordering penalties.  Illustratively, we depict the useful dimensions of the input in 'red' color (or its variants). The idea is that the representation $\mU$ can filter this dimension via contrasting against the noise.}
    \label{fig:architecture}
\end{figure*}

Suppose we are given a set of negative examples to work with. As the performance of contrastive learning depends on the coupling between positive and negative examples, to this end, we propose a novel contrastive learning objective that \emph{simultaneously optimizes} for learning a representation via a tightly coupled interplay between four key components, namely (i) generating a (probabilistic/soft) coupling with the negative data via \emph{minimizing} the optimal transport /Wasserstein distance \cite{OTAM_book} between the positive and negative examples, (ii) a representation learning cost that seeks a low-dimensional data subspace such that the projections of the data onto this subspace \emph{maximizes} their distance from the coupled negative examples, (iii) a \emph{distortion} penalty that prevents the subspace projections from diverging too much from the input data, and (iv) an ordering constraint on the subspace projections that captures the sequentiality of the input. 

Having outlined our novel approach for simultaneous learning of coupling between the positive and negative examples and the (contrastive) representation, we turn towards addressing generation of good negative examples. As pointed out in \cite{Tschannen2019}, design of good negative distribution and its impact on contrastive learning is not fully addressed yet. In contrast to existing works such as, \cite{Henaff2019,oord2018representation,Sun2019,WangG15a} we show that, when coupled with the objective to learn a contrastive representation, one must allow a high discrimination, with respect to the intended task, from the positive distribution while maintaining good correlation. 

We resolve these conflicting objectives in learning the contrastive distribution by trading off conditional distributional discrepancy with a (given) classifier accuracy on the contrastive distribution. Specifically, we use a Wasserstein GAN that takes as input positive examples from the dataset, and learn to generate new samples (conditioned on the positives), where these generated samples when added to their respective positive examples will ensure two properties, namely (i) the modified positives must be as close in distribution as possible to the true positives in the dataset, and (ii) a classifier that has good performance on the true positives must have high misclassification rate on these modified positives. Figure~\ref{fig:architecture} pictorially depicts our complete framework. 

We summarize our novel contributions below.
\begin{enumerate}
\vspace*{-2pt}
    \item We propose a representation learning cost that \emph{naturally} blends contrastive learning, adversarial learning, optimal transport, and Riemannian geometry into one framework.
    \item We show that our objective for learning contrastive representation, while completely differing in its aims, is related to the subspace robust optimal transport distances proposed in \cite{Paty_Cuturi2019}. We characterize this relation in Theorem~\ref{thm:srot_cot}, thereby making a novel connection between contrastive learning and robust optimal transport.
    \item Further, we present an adversarial distribution learning setup within which, it is possible to optimize over and regulate the choice of distribution for negative samples that is known to be critical for contrastive representation learning.
    \item We apply this framework for the problem of learning representations for classifying action video sequences and obtain promising classification results.
\end{enumerate}

\section{Related Work}
\label{sec:related_work}
Our approach has several parallels and similarities with existing works that we systematically outline below. Contrastive estimation~\cite{smith2005contrastive} and noise-contrastive estimation~\cite{gutmann2010noise} are popular representation learning methods that learn via an objective that contrasts data likelihood under the model against the likelihood of the noise or implicitly-constructed contrastive (i.e., negative) examples. Popular deep metric learning and triplet losses~\cite{hoffer2015deep} can also be considered as variants of contrastive learning when explicit pairwise relationships between data samples are available. As empirically shown in many works and recently theoretically argued in~\cite{arora2019theoretical}, contrastive learning can reduce the sample complexity for downstream tasks. 

There have been recent efforts at unifying the general representation learning methods and contrastive methods ~\cite{hoffer2015deep, arora2019theoretical}, such as for example via the maximum InfoNCE principle \cite{Tschannen2019, oord2018representation}, in which the main idea is to implicitly maximize the mutual information~\cite{cover2006} between the learned representation and the original data. However, these formulations assume a good approximation of mutual information from the given samples -- a problem that is typically hard. Recently, using different forms of variational characterizations of mutual information \cite{pmlr-v97-poole19a}, efficient estimators have been proposed. Nevertheless, the resulting optimization and training may become unstable~\cite{song2019understanding}.

Among works based on maximizing InfoNCE, one similar to ours is~\cite{oord2018representation} that proposes contrastive predictive coding through a density ratio capturing the mutual information between the representation and future samples of the sequence. They use noise contrastive estimation on their representation learning cost, where the noise distribution is considered as those plausibly unrelated to the input. Our proposed framework is different from theirs in that we explicitly seek a negative distribution that can potentially increase the contrastiveness of useful cues via learning to generate these hard negative samples. Attempts towards improving the negative sampler in contrastive learning have been made in \cite{Avishek_ACE}. Here the authors propose to use a mixture of unconditional and conditional negative distributions, conditioned on given data, and parametrized via an implicit generative model; however with a totally different loss function than ours. In contrast, we achieve the required discrimination using a classifier, while our proposed variant of WGAN captures the similarity of the negatives to the given data.

Our work is also related to discriminative pooling~\cite{Wang_Cherian_TPAMI2019} that proposes to generate negative samples via passing random noise through an image-trained CNN, however is not adversarial. In~\cite{Wang_Cherian2018}, the authors propose an adversarial setup in a discriminative representation learning framework, however uses a deterministic deep model to learn a single adversarial sample per data point. Instead, our proposed framework can generate distributions of adversarial samples. Another paper related to ours is~\cite{cherian2017generalized} that proposes to use subspace representations for video sequences. While, we also use their temporal learning constraints, our optimal transport and adversarial distribution learning offer a richer representation learning setup via suppressing perhaps false-positive temporal features, such as temporally-evolving noise.

\section{Problem Formulation}
\label{sec:background}
In this section, we describe our problem setup, introduce our notation, and review some prior work on which we build our proposed algorithms. Following standard notation, we use uppercase boldface letters  $\M{X}$ to denote matrices, and lowercase boldface $\vx$ to denote vectors. Refer Figure~\ref{fig:architecture} for contextualizing our notation and variables in the sequel.

Suppose we are given a set of $N$  data sequences $\dataset=\set{\mX_1, \mX_2,\cdots, \mX_N}$, where each $\mX_i=\seq{\vx^i_1,\vx^i_2,\cdots, \vx^i_{n_i}}$ is a sequence of $n_i$ ordered feature vectors and each $\vx_t\in\reals{d}$. Further, we assume $\mX_i$ is associated with a ground truth class label $\ell_i\in\labels$; $\labels$ denoting a given set of labels. We also assume each $\vx\in\mX$ is an independent sample from a data distribution $\pD(\bmX)$ conditioned on the mean $\bmX$ of the sequence $\mX$.\footnote{We may use any other central tendency of the sequence to define this distribution.} Note that we do not make any explicit assumption on what these sequences represent. For example, in a video recognition application, each $\vx_t$ could be the output of a frame-level deep neural network that is trained on individual frames against their respective video label~\cite{simonyan2014two}.

As each feature $\vx_t$ in a sequence $\mX$ is assumed to be generated independently without accessing the rest of the sequence, these individual features could be noisy; for example, they could be entangled with irrelevant features from the background. Our key assumption is that the \emph{useful} temporally-evolving features belong to subspaces in this $d$-dimensional feature space. Thus, our main idea is to design an objective that could extract these subspaces in a compact form, denoted $\mU(\mX)$, such that by using this representation some suitable empirical sequence recognition loss $\dataloss$ is minimized, where:
\vspace*{-3pt}
\begin{equation}
    \dataloss := \frac{1}{N}\sum_{i=1}^N \vidloss(\mU_i, \ell_i) \text{ and } \mU_i = \argmin_{\mU} \reploss(\mU(\mX_i)).
    \label{eq:classifier}
\end{equation}
The loss $\dataloss$ aggregates the error $\vidloss$ in training a classifier $\classifier$ on the representations $\mU_i$ for each sequence against its ground truth label $\ell_i$. Further, the $\mU_i$'s are obtained via optimizing a sequence level representation learning objective captured by the loss $\reploss$. In a classical feature learning setup~\cite{simonyan2014two,carreira2017quo}, $\reploss$ finds a vector $\mU$ that minimizes, say, the mean-squared error to the data samples; which boils down to the average feature. Thus, in that case, the $\argmin$ optimization is merely the average pooling scheme. In this paper, we generalize this pooling for richer and better representation learning.  While, we can easily train for the two losses $\vidloss$ and $\reploss$ jointly in an end-to-end manner~\cite{Wang_Cherian_TPAMI2019}, in this work, we deal with them separately so that we have better control of each of them. In the next few sections, we look deeper into the representation loss using a contrastive learning framework. We will describe the classifier loss $\vidloss$ in Sec.~\ref{sec:classifier}

\subsection{Contrastive Learning via Optimal Transport}
\label{sec:contrastive}
Suppose, we treat a sequence $\mX_i$ as a \emph{set} of positive examples (ignoring the order within), and assume that we have access to a set of negative examples, denoted $\mY_i=\set{\vy^i_1,\vy^i_2,\cdots, \vy^i_m}$, each $\vy\in\reals{d}\sim\prob{\mY}$. In noise contrastive estimation, $\pY$ is typically assumed to be either uniform or Gaussian noise. The goal of contrastive learning in this setup is to find a suitable representation for $\mX$ that is maximally ``distant'' from $\mY$. How should we characterize this contrastiveness? Given that we are working with sets of data points under the assumption that they are random samples from underlying probability distributions, a natural possibility is to consider the Optimal Transport (OT), also known as the Wasserstein distance between the two distributions \cite{OTAM_book}.

Recall that the Wasserstein distance, denoted by $W_c(\mu, \nu)$, between two probability measures $\mu$, and $\nu$ that are both supported on $\mb{R}^d$ with respect to a ground cost $c(\V{x}, \V{y}), \V{x}, \V{y} \in \mb{R}^d$, is given by \cite{OTAM_book}: 
 \begin{equation}
     W_c(\mu, \nu) := \inf_{\pi\in\Pi(\mu, \nu)}\!\!\!\!\expect{(\vx,\vy)\sim\pi}\ c(\vx, \vy),
     \label{eq:wasser}
 \end{equation}
 where $\Pi(\mu,\nu)$ denotes the set of all couplings (joint probability distributions) with marginals $\mu$ and $\nu$. Typically, in the absence of any other information and when the measures are discrete, $\mu$, $\nu$ are chosen to be uniform distributions over the support. Specific to our case, given positive examples $\M{X}$, we let $\mu_{\M{X}}$ be the empirical distribution (with equal, i.e., uniform probability) over $\V{x}_t \in \M{X}$, i.e., $\mu_\M{X} = \sum_{t =1}^{n} \frac{1}{n} \delta(\V{x}_t)$, $\delta(\V{x}_t)$ denoting the Dirac measure at $\V{x}_t$. Similarly, let $\nu_\M{Y}$ be the empirical distribution of the negative samples $\M{Y}$, also uniform over the samples.  
 
 Next, let $\fU:\mb{R}^d \rightarrow \mb{R}^d$ be a mapping parameterized by the \emph{to be learned} representation $\mU$. Then, we formulate our constrastive optimal transport problem for representation learning as:
 \begin{equation}
    \vspace*{-3pt}
     \max_{\mU}\ \lot(\mU):= W_c({\fU}_{\#}\mu_{\mX}, \nu_{\mY}).
     \label{eq:ot}
 \end{equation}
   The notation ${\fU}_\# \rho$ denotes the push-forward measure of $\rho$ under the mapping $\fU$. In this paper, we assume the useful features belong to linear feature subspaces\footnote{This \emph{linearity} choice is inspired by the observation that usually these deep features are extracted from the penultimate layers of a CNN, subsequently classified using a linear classifier.} and thus use the mapping $f$ defined as $f=\mU\mU^{\top}$ for orthonormal $\mU\in\reals{d\times k}$, i.e., $\mU^{\top}\mU=\M{I}_k$, where $\M{I}_k$ denotes the $k\times k$ identity matrix and $k \ll d$. Rewriting the Wasserstein distance~\eqref{eq:wasser} in empirical form, combining it with the definition of $f$ in~\eqref{eq:ot}, and using the $\ell_2$-norm for the OT cost $c$, we can write the contrastive representation learning objective as:
   \begin{equation}
       \max_{\mU\in\grassmann(d,k)} \lot(\mU) :=  \inf_{\pi\in\Pi(\mu_{\mX}, \nu_{\mY})} \sum_{i,j}\pi_{ij} \|f_\M{U}(\V{x}_i) - \V{y}_j\|,%\enorm{\mU\mU^{\top} \mX - \mY\pi}.
       \label{eq:cot}
   \end{equation}
In the formulation~\eqref{eq:cot}, we assume $\mU\in\grassmann(d,k)$, the Grassmann manifold of all $k$-dimensional subspaces of $\reals{d}$. Recall that $\grassmann(d,k)$ denotes the quotient space $\cS(d,k)/\cO(k)$ of all $d\times k$ orthonormal matrices $\cS(d,k)$ that are invariant to right rotations. Given that our loss $\lot(\mU)=\lot(\mU R)$ for any $k\times k$ orthogonal matrix $R$, casting the learning objective on the Grassmann manifold is a natural choice. 

A question with regard to~\eqref{eq:cot} is why we have $\fU$ acting only on the positive samples? This is because, we assume that the negative samples (as described in Sec.~\ref{sec:noise}) share all the subspaces of the positives, except for those subspaces containing relevant cues for classification (e.g., action-related), which are present only in the positives. Thus, asking for a maximal common projection subspace on positive and negatives may lead the optimizer to move away from finding the useful contrastive subspaces in the positives. % 

\subsubsection{Connections to Subspace Robust OT}
We note that \eqref{eq:cot} is itself novel for contrastive learning in that instead of looking for pairs of positive and negative examples as is common in contrastive learning setup, it implicitly learns the best coupling (via the optimal transport plan) and enforces this coupling to be the maximal. Our formulation~\eqref{eq:cot} is related to the recently proposed subspace robust Wasserstein distances \cite{Paty_Cuturi2019}, which can be defined in our problem setup as:

\begin{align*}
    \mc{P}^2_k &\triangleq  \max_{\M{U}: \cS(d,k)}    \min_{\pi\in \Pi(\mu_{\mX}, \nu_{\mY)})} \mb{E}_{\pi} \| \M{U}^\top \V{x} - \M{U}^\top \V{y}\|^2,\text{ and } \\\mc{S}_k^2 & \triangleq \min_{\pi\in\Pi(\mu_{\mX}, \nu_{\mY})} \max_{\mU\in\grassmann(d,k)} \mb{E}_{\pi} \| \M{U}^\top \V{x} - \M{U}^\top \V{y}\|^2.
\end{align*}
Suppose, $\mc{C}^2_k$ denotes the Wasserstein-2 distance variant of our objective in~\eqref{eq:cot}; i.e.,
\begin{equation}
    \mc{C}^2_k = \max_{\mU\in\grassmann(d,k)} \min_{\pi\in\Pi(\mu_{\mX}, \nu_{\mY})} \sum_{i,j}\pi_{ij} \|f_\M{U}(\V{x}_i) - \V{y}_j\|^2.
\end{equation}
It is clear that our subspaces $\mU$ act only on the positive examples, and seek the maximal robustness against the chosen negative distribution. The following theorem characterizes the connection between the solutions to the two objectives. The proof can be found in the supplementary material.

\begin{thm}
\label{thm:srot_cot}
Assuming $n_Y$ negative samples, 
\begin{align} &\mc{P}_k^2 \leq \mc{C}_k^2 \leq  \mc{S}_k^2+ \max_{\mU\in\grassmann(d,k)} \frac{1}{n_Y}\!\!\sum_{j=1}^{n_Y} \|(\M{I}_d - \mU\mU^{\top}) \V{y}_j \|^2, \notag
\end{align} with equality iff $k = d$. The variational term on the RHS is equal to the sum of $d-k$ largest eigenvalues of the Gram matrix $\M{\Sigma}_\M{Y} = \frac{1}{n_Y} \sum \V{y}_j \V{y}_j^\top$. % and can be estimated from the data. 
\end{thm}

\subsection{Generating Noise Distributions}
\label{sec:noise}
As alluded to above, in contrastive learning, typically the noise distribution is assumed uncorrelated to the data. However, it is often observed that as the noise is closer to the data (hard negatives), the quality of the learned representation improves. In this regard, there are two difficulties to address with regard to the noise generation, (i) how to generate the noise distribution $\nu_Y$ closer to the data distribution $\mu_X$, and (ii) how to ensure the generated noise will not impact the useful features in $\mX$? We resolve this dilemma via generative adversarial networks with some new regularizations. 

Concretely, suppose our measure on the negative samples $\nuY$ is defined via an implicit function $\gth:\reals{d}\to\reals{d}$, where $\theta$ defines its parameters to be learned. That is, we assume $\vy=\gth(\vz)$, where $\vz\sim\normal(\bmX, \sigma^2\eye)$. Specifically, we assume the input to our implicit generator $\gth$ comes from \emph{normally} distributed data with mean $\bmX$ and standard deviation $\sigma$. Note that $\bmX$ defines the average of the samples in the respective sequence, i.e., $\bmX=\frac{1}{n}\sum_{t}\vx_t$. Recall that we had originally assumed each $\vx\in\mX\sim\pD(\bmX)$ (in Sec 3). Now, our goal is to learn $\theta$ such that it emulates $\pD(\bmX),\ \forall \mX\in\dataset$, while also capturing other desirable properties listed above. 
Substituting the definition of $\vy$ in~\eqref{eq:cot}, we have a modified OT problem:
\begin{align}
    \lot':= \min_{\theta} \min_{\pi\in\Pi(\mu_{\mX},\gth(\bmX,\sigma))} \expect{\pi} \enorm{\fU(\vx) - \vy},
    \label{eq:implicit}
\end{align}
where we use the succinct notation $\gth(\bmX,\sigma)$ to explicitly show the relation of the noise distribution $\nu_Y$ with the input sequence $\mX$. Using Wasserstein-1 distance, we can use the Kantorovich duality to derive a dual form of this objective via learning a 1-Lipschitz function $h$, and rewriting~\eqref{eq:implicit} as:
\begin{equation}
    \lot'':=\min_{\theta} \max_{h\in L_1}  \expect{\vx\sim\mu_{\mX}} \sqbrack{h(\fU(\vx))} - \expect{\vy\sim\gth(\bmX,\sigma)} \sqbrack{h(\vy)}.
    \label{eq:wgan-ot}
\end{equation} 

The formulation in~\eqref{eq:wgan-ot} is a variant of the popular Wasserstein-GAN~\cite{arjovsky2017wasserstein}, except that the noise $\vz$ is conditioned on the input sequence, and that the input data $\vx$ is projected into some subspace $\mU$ to be learned. 

Note that while it may seem we can optimize all the parameters together alongside learning the representation $\mU$ using \eqref{eq:wgan-ot}, it poses some technical hurdles. For example, unless we know how to generate the negative distribution via learning $\theta$, we cannot produce the representation $\mU$, and without this representation, we cannot use $\fU$. This poses a challenge with learning the two jointly, especially when considering highly non-linear neural networks to characterize $h$. A second problem is that the data from a single sequence may be insufficient to learn $\theta$. We circumvent both these problems by separating the representation learning objective from the noise generation objective; the latter we re-formulate as:
\begin{equation}
    \lgan(\theta) :=  \min_{\theta} \max_{h\in L_1} \expect{\vx\in\mX\sim\dataset} \sqbrack{h(\vx)} - \expect{\vy\sim\gth(\bmX,\sigma)} \sqbrack{h(\vy)},
    \label{eq:new_wgan}
\end{equation}
where now, we removed $\fU$ and included the parameter learning problem using the full dataset $\dataset$, instead of a single sequence $\mX$. 

\subsection{Adversarial Noise Generation}
While, our formulation in~\eqref{eq:new_wgan} does allow learning noise distributions that are similar in distribution to the data samples $\vx$, it is not adversarial in the sense that there is nothing preventing the generated noise from being adequately discriminative, i.e., from mimicking spatio-temporal features; for example, $\mY$ may have features that are useful for recognition, however we desire our final representation to be maximally different from this noise. To account for these requirements, we propose to learn to generate adversarial noise. We need some new notation to present our technique.

Suppose we have a pre-trained (frame-level) classifier $\fcl:\reals{d}\to|\labels|$ that takes each $\vx\in\mX$ and returns a class label $\ell_{\mX}$ associated with $\mX$. Different from~\eqref{eq:wgan-ot}, now our objective is not to just produce noise, but to make this noise adversarial to the useful components in the input data. That is, for a given feature $\vx\in\mX$, we seek to generate a noise sample $\hat{\vx}$ from $\gth(\bmX,\sigma)$ such that when this sample is subtracted from the input $\vx$, a classifier $\fcl(\vx)$ that is trained to produce $\ell_{\mX}$ will misclassify; i.e., if $\fcl(\vx)\to \ell_{\mX}$, then
$\fcl(\gamma(\vx-\hat{\vx}))|\hat{\vx}\sim\gth(\bmX,\sigma)\to \bar{\ell}_{\mX}$
, where $\bar{\ell}_{\mX}\in\labels$ is any other class label\footnote{In practice, WGAN usually learns to generate random noise of arbitrary strength that could misclassify the input, without accounting for useful subspaces. To circumvent this issue, our objective demands that the generated noise when combined with the input data will ensure the useful data properties are removed. We do this by asking the classifier to produce a logit vector such that its softmin is equal to $\ell$, while for the non-perturbed input, we ask the softmax be equal to $\ell$.} other than $\ell_{\mX}$, and $\gamma$ is a suitable operator; e.g., a ReLU ensuring $\gamma(\vx-\hat{\vx})$ remains in the same feature space as $\vx$. Incorporating these requirements into our GAN objective in~\eqref{eq:wgan-ot}, we have our new adversarial WGAN objective as:
\begin{align}
    &\ladv(\theta) =  \min_{\theta} \max_{h\in L_1} \expect{\vx\in\mX\sim\dataset} \sqbrack{h(\vx)} - \expect{\substack{\vy=\gamma(\vx-\hat{\vx}),\\\hat{\vx}\sim\gth(\bmX, \sigma)}} \notag \sqbrack{h(\vy)} \\
    &\quad+ \lambda_1\left(\fcl(\vx, \ell_{\mX}) - \fcl(\gamma(\vx-\hat{\vx}),\ell_{\mX})\right)\!+\!\lambda_2 \mb{E}[\enorm{\hat{\vx}}^2].
    \label{eq:adv_wgan}
\end{align}
The regularization $\mb{E}[\enorm{\hat{\vx}}^2]$ is useful to make $\gth$ learn to produce noise of small strength that can misclassify the input (similar to standard adversarial models~\cite{moosavi2016deepfool}). The positive weights $\lambda$'s balance the losses at training. Note that we need to input $\bmX$ to the generator $\gth$ so that the generator learns to know the noisy features in its input (at the frame level). Our full objective in~\eqref{eq:adv_wgan} is trained end-to-end for the WGAN.

\subsection{Capturing Sequentiality and Distortion }
Now that, we have a concrete setup to sample from adversarial noise distributions to generate the negative samples, lets look back at the representation learning objective in~\eqref{eq:cot}. Recall that using the noise distributions, we have accounted for finding samples $\mY$ that do not have \emph{useful} data properties that were present in the input data $\mX$. While, using $\mY$ in OT allows for a high and relevant discrimination from $\mX$; however, the constrastive representation should also account for the sequentiality in the data. To this end, we include temporal ordering constraints on the learned subspaces. Specifically, we ask the subspace to be learned in such a way that when data points is projected onto this subspace, some monotonicity property is satisfied. That is, for some $\eta>0$, the projections of each $\vx_t$ onto the learned subspace $\mU$ satisfies ordering constraints: $\enorm{\mU^{\top}\vx_t}^2+\eta\leq \enorm{\mU^{\top}\vx_{t+1}}^2, \forall t=1,2,\cdots, n-1$. We also ensure that the representation does not diverge too much from the input sequence, via including a distortion penalty into our objective. Note that these constraints have been used previously, such as in~\cite{cherian2017generalized}. With these additional constraints, we rewrite our full representation learning objective in~\eqref{eq:cot} as:
\begin{align}
    \max_{\mU\in\grassmann(d,k)} &\reploss(\mU) :=  \lot(\mU) - 
    \frac{\beta_1}{n}\sum_{\vx\in\mX}\enorm{\fU(\vx)-\vx}^2 - \notag\\ &\frac{\beta_2}{n-1}\sum_{t=1}^{n-1}\hinge{\enorm{\mU^{\top}\vx_t}^2 +\eta - \enorm{\mU^{\top}\vx_{t+1}}^2},
    \label{eq:full-objective}
\end{align}
where the $\beta$s are constants, and $\hinge{\ .\ }=\max(0,.)$. 

\subsection{Representation Learning}
For a given sequence $\mX\in\reals{d\times n}$ and its adverserially sampled noise matrix $\mY\in\reals{d\times m}$, we can rewrite our full objective in~\eqref{eq:full-objective} as:
\begin{equation}
    \max_{\mU\in\grassmann(d,k)}\min_{\pi\in\Delta(n, m)} \inner{\pi, \dist(\fU(\mX), \mY)} - \Omega(\mX, \mU), 
\end{equation}
where $\Delta(n, m)$ is a set of linear constraints capturing the marginals, and $\dist$ produces an $n\times m$ distance matrix. Further, with a slight abuse of notation, we assume $\fU(\mX)=\mU\mU^{\top}\mX$, and $\Omega(\mX, \mU)$ captures the distortion and the ordering constraints on $\mU$. We propose to use alternating minimization to solve this objective, where we alternatingly solve the following sub-problems, while keeping the other optimization variable fixed. Specifically, by fixing $\mU$ (initially assuming $\mU\mU^{\top}=\eye$), we solve for $\pi$ as:
\begin{equation}
    \min_{\pi\in\Delta(n\times m)} \inner{\pi, \dist(\fU(\mX, \mY)},
\end{equation}
which we solve efficiently using  an inexact proximal point optimal transport (IPOT) solver proposed in \cite{ipot}. In contrast to entropy regularization based methods \cite{Cuturi_2013}, IPOT has better convergence and stability properties. 
Fixing the coupling $\pi$, we solve for the subspace $\mU$ via casting the optimization on the Grassmann manifold. That is, we solve:
\begin{align}
    \min_{\mU\in\grassmann(d,k)} -\fnorm{\mU\mU^{\top}\mX - \mY\pi^{\top}}^2 + \Omega(\mX, \mU).
\end{align}
We use the Riemannian conjugate gradient algorithm~\cite{absil2009optimization} to optimize this sub-problem. We use the Fletcher and Reeves step size selection~\cite{fletcher1964function} and retractions via the QR decomposition for the iterations. As each of our sub-problems is non-convex, it is not easy to guarantee any convergence. However, empirically, we see that the IPOT solver finds the coupling in about 1000 iterations (note that each iteration is very cheap) and the Riemannian conjugate gradient converges in about 5 iterations.

\subsection{Representation Classifier}
\label{sec:classifier}
The missing piece to present in our framework is the classifier $\classifier$ to use on the subspace representations, defined in~\eqref{eq:classifier}. A tricky problem with subspaces is that there is no control on the sign of their basis.  While, we may use a deep neural network classifier to this end, in this work, we use a standard kernelized SVM, with a kernel $\kernel$ defined for two points $\mU_1,\mU_2\in\grassmann(d,k)$ as
~\cite{harandi2014expanding}:
\begin{equation}
    \kernel(\mU_1, \mU_2) = \exp\left(\gamma \fnorm{\mU_1^{\top}\mU_2}^2\right),
\end{equation}
for a bandwidth parameter $\gamma>0$.  A computationally cheaper alternative to using a kernel, which we found to produce similar results empirically, is to use average pooling after computing $\mU\mU^{\top}\mX$. This results in a representation in $\reals{d}$, and is especially desirable if using a linear sequence classifier.

\begin{table*}[ht]
\begin{center}
\begin{tabular}{l|ccc|ccc|ccc}
    \hline
    \multicolumn{1}{l|}{Ablation}&\multicolumn{3}{c|}{JHMDB (vgg)} & \multicolumn{3}{c}{JHMDB (I3D)} & \multicolumn{3}{|c}{HMDB (I3D)}\\
     & RGB & FLOW & R+F & RGB & FLOW & R+F & RGB & FLOW & R+F\\
    \hline
    Avg. Pool & 47.0 & 63.0 & 73.1 & 77.5 & 81.0 & 85.0 & 68.2 & 69.5 & 76.5\\
    \hline  
    COT + Random & 48.0 & 63.9 & 77.9 & 62.2 & 77.2 & 79.4 & 68.5 & 71.1 & 72.5\\
    ACOT & 49.3 & 65.0 & 75.0 & 76.1 & 81.2 & 90.0 & 69.5 & 74.6 & 76.4\\
    ACOT + PCA & 49.5 & 65.7 & 75.6 & 77.6 & 82.8 & 90.6 & 69.8 & 74.9 & 76.6\\
    AC + PCA + order (No OT) & 49.0 & 66.1 & 75.8 & 75.2 & 80.0 & 89.8 & 70.2 & 74.8 & 76.3\\
    ACOT + PCA + order & \textbf{50.3} & \textbf{69.2} & \textbf{79.8} & \textbf{78.1} & \textbf{82.9} & \textbf{91.5} & \textbf{73.2} & \textbf{75.5} & \textbf{79.4}\\
\end{tabular}
\end{center}
\caption{Ablative Study of various modules in our framework and their benefits on the JHMDB dataset (with two different types of frame-level features, i.e.,VGG and I3D) and the HMDB dataset. We report performances on these datasets for the RGB stream, the optical flow stream, and their combination in a two-stream action recognition setup. The results are based on the split-1 of the respective datasets. The acronyns are as follows: A=Adversarial, C=Contrastive, OT=optimal transport, PCA=principal components analysis, and Random=using random noise instead of adversarially generated noise, order=Temporal ordering. Note that for this experiment, we did not fine-tune the I3D features on the datasets; instead use the pre-trained Imagenet+Kinetics model~\cite{carreira2017quo}.}
\label{tab:ablative}
\end{table*}
\begin{figure*}[ht]
    \centering
    \subfigure[Increasing subspaces $k$]{\label{fig:subspaces}\includegraphics[width=4.2cm,trim=0.2cm 6cm 0.2cm 5cm,clip]{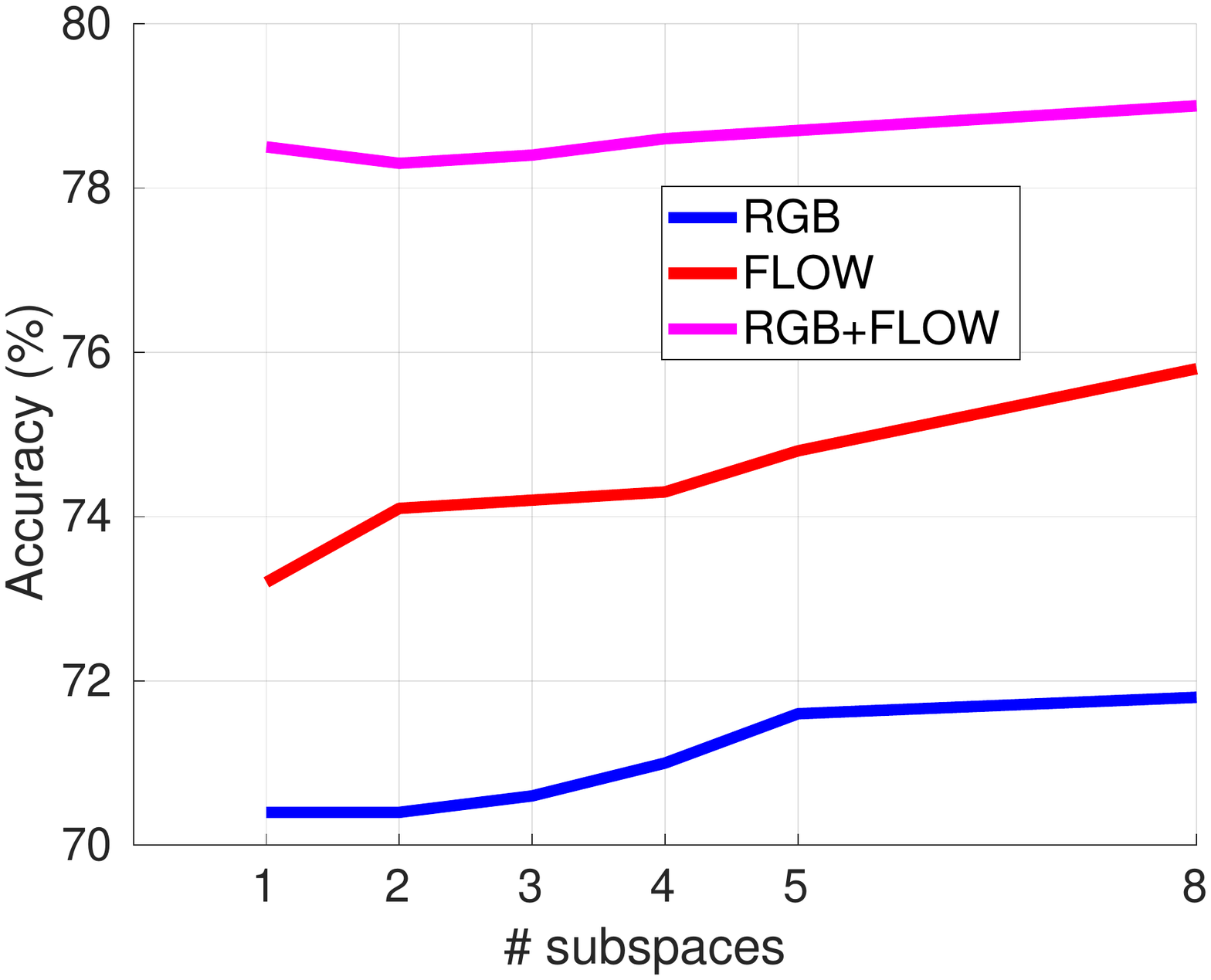}}
    \subfigure[Distortion regularization $\beta_1$]{\label{fig:pca_val}\includegraphics[width=4.2cm,trim=0.2cm 5.5cm 0.2cm 5cm,clip]{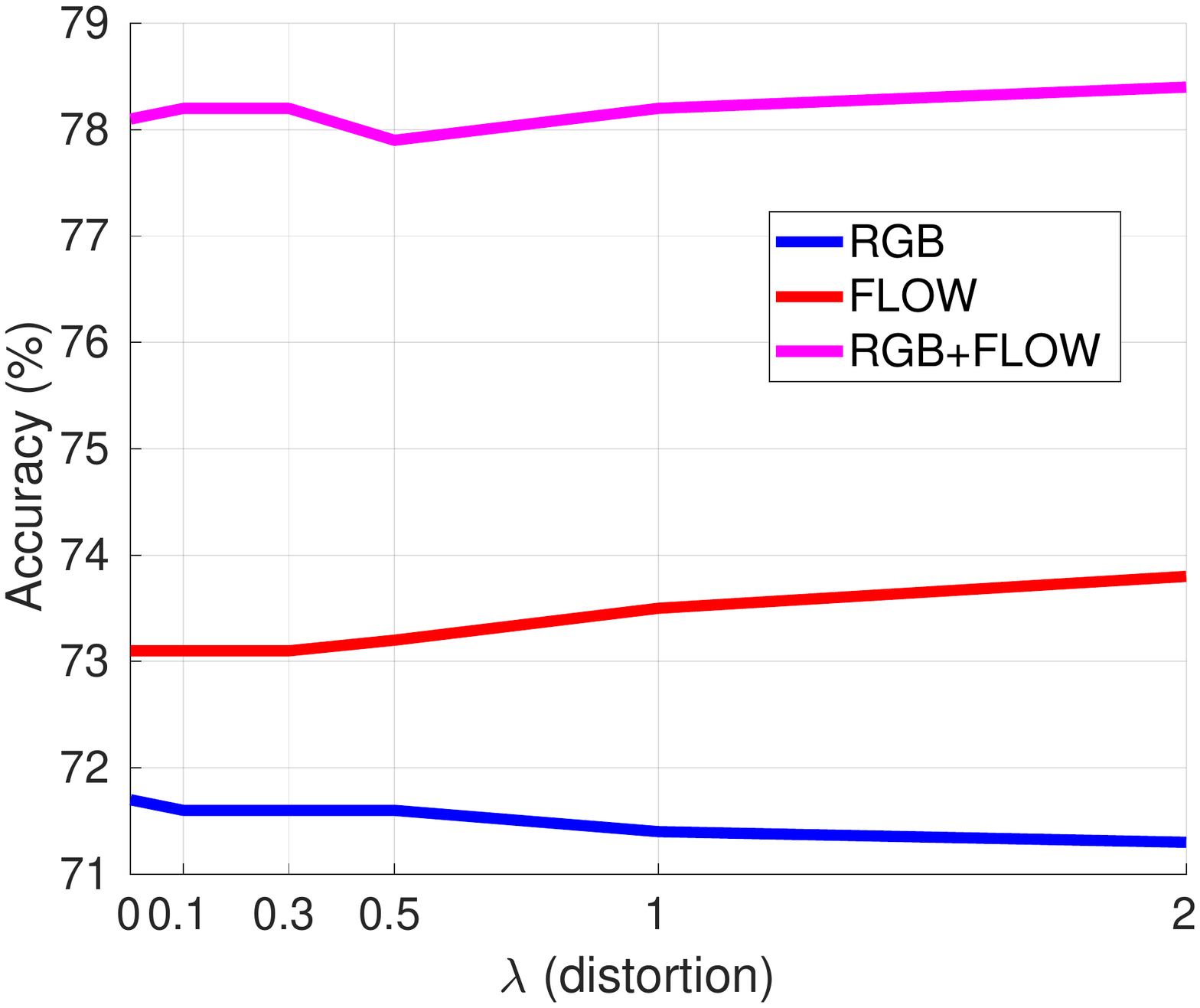}}
    \subfigure[$\#$ Negatives]{\label{fig:num_negs}\includegraphics[width=4.2cm,trim=0.2cm 6cm 0.2cm 5cm,clip]{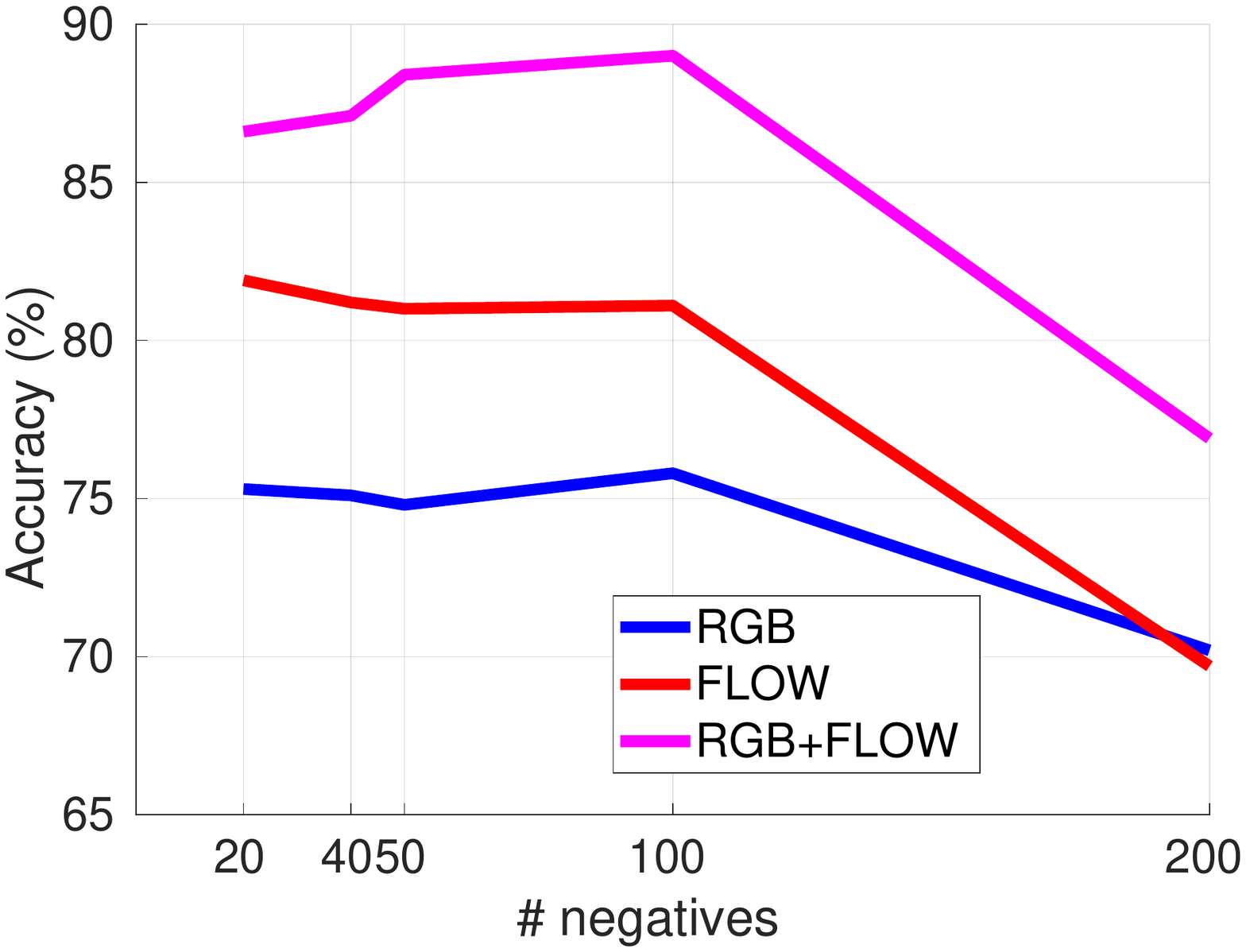}}
    \subfigure[Temporal regularization $\beta_2$]{\label{fig:beta2}\includegraphics[width=4.2cm,trim=0.2cm 6cm 0.2cm 5cm,clip]{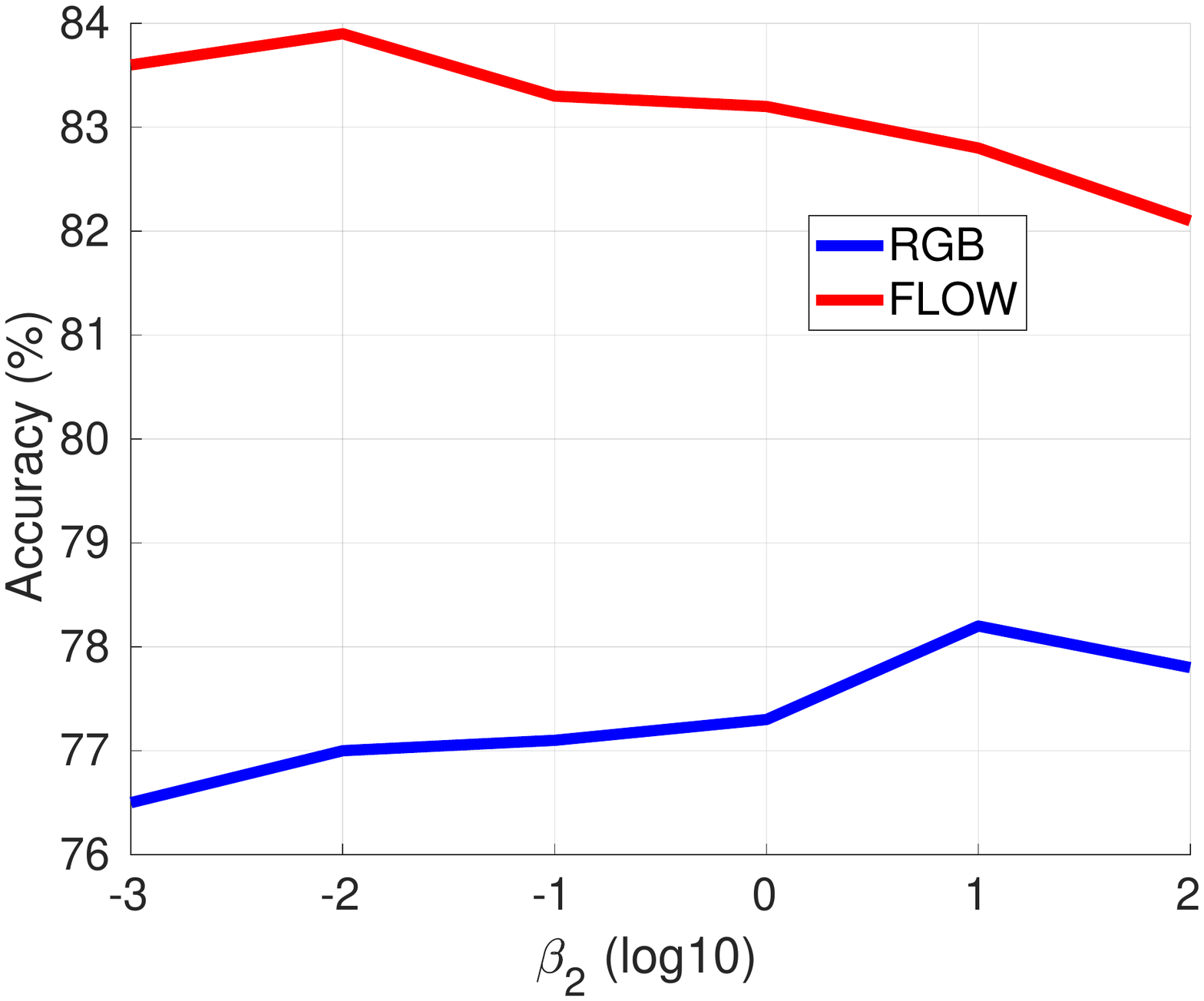}}
    \caption{Sensitivity analysis of various choices in our representation learning setup. See text for details. Left two plots are on the HMDB dataset, the right two are on the JHMDB dataset.}
    \label{fig:ablative}
\end{figure*}

\section{Experiments}
\label{sec:expts}
In this section, we present experiments demonstrating the effectiveness of our approach for representation learning. To this end, we use two standard action recognition datasets, namely (i) small-scale JHMDB~\cite{jhuang2013towards} and (ii) the larger HMDB dataset~\cite{kuehne2011hmdb}. We also present some qualitative results on the CIFAR dataset. More experiments and results are provided in the supplementary materials, due to lack of space.

\noindent\textbf{JHMDB dataset:} consists of 928 video sequences, each sequence about 15-40 frames long. There are 21 actions defined on the clips, and each clip has only one action. We use this dataset as a test-bed to explore the performances of various modules in our setup. To this end, we use a standard two stream neural network~\cite{simonyan2014two} for extracting video features at the frame-level and from a short-clip of 20 optical flow frames. To make our results comparable to prior works, we use vgg-16 features made available as part of~\cite{cherian2017generalized}. We also evaluate using 3D CNN features as produced by a pre-trained I3D network~\cite{carreira2017quo} in a two-stream setup.

\noindent\textbf{HMDB Dataset:} is a super-set of the JHMDB dataset and consists of about 6700 video clips, each with 50--400 frames and 51 actions. We use a pre-trained I3D network (trained on Kinetics) in a two-stream framework for evaluating the performance of our model on this dataset.

\noindent\textbf{Hyperparameters:} Our entire implementation is in PyTorch. For our adversarial module, we modified the public WGAN code associated with~\cite{arjovsky2017wasserstein}. We used a noise variance $\sigma=0.01$, which resulted in an average classifier fooling rate of 60\% on the training set on both the datasets. See the Appendix for more experiments in this regard.  Further, we used $\lambda_1=0.1$ and $\lambda_2=1$ in~\eqref{eq:adv_wgan}. We used PyManOpt as our Riemannian optimization framework\footnote{\url{https://www.pymanopt.org/}}. As for the regularization constants on the distortion and ordering constraints in~\eqref{eq:full-objective}, we set $\beta_1=1$ and $\beta_2=10$, and we used $\eta=0.01$ for the temporal margin. We also assume that all features from the two datasets are normalized to unit $\ell_2$ norm.  We will be making our code publicly available at \url{https://www.merl.com/research/license/}.

\begin{figure*}[ht]
    \centering
    \includegraphics[width=5.5cm]{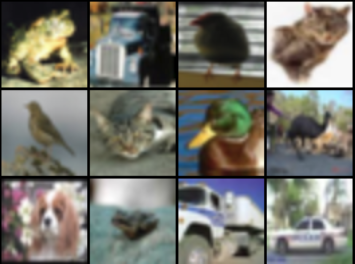}
    \includegraphics[width=5.5cm]{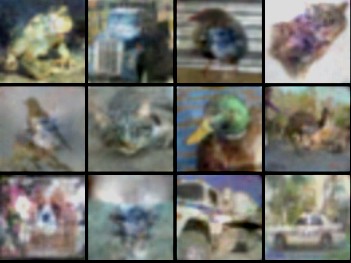}
    \includegraphics[width=5.5cm]{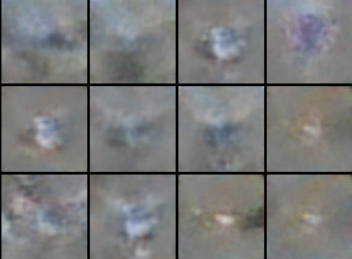}
    \caption{Qualitative results showing (left) the original CIFAR image, (middle) image added with sampled noise from an adversarial distribution, and (right) the respective sampled noise. See text for details.}
    \label{fig:cifar}
\end{figure*}
\begin{figure*}[ht]
\center
\subfigure[Adv. Samples]{\label{fig:tsne-a}\includegraphics[width=4.2cm,trim={0.5cm 0.5cm 0.5cm 0.5cm}, clip]{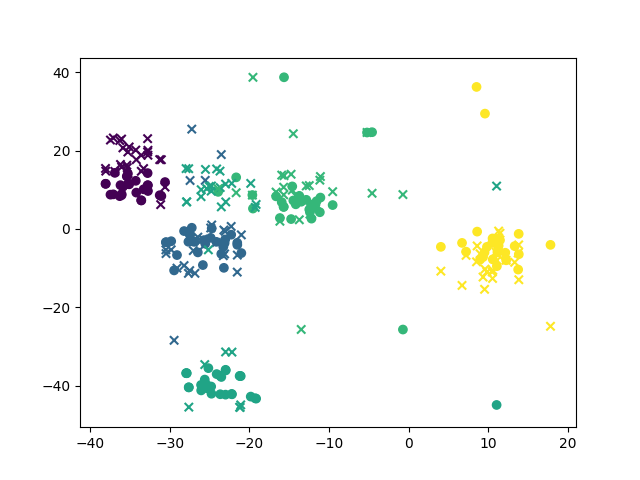}}
\subfigure[Average Pooling]{\label{fig:tsne-b}\includegraphics[width=4.2cm,trim={0.5cm 0.5cm 0.5cm 0.5cm}, clip]{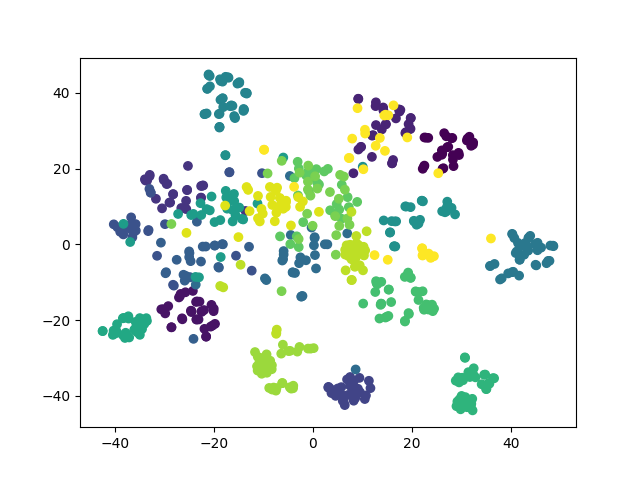}}
\subfigure[Random Contrastive]{\label{fig:tsne-c}\includegraphics[width=4.2cm,trim={0.5cm 0.5cm 0.5cm 0.5cm}, clip]{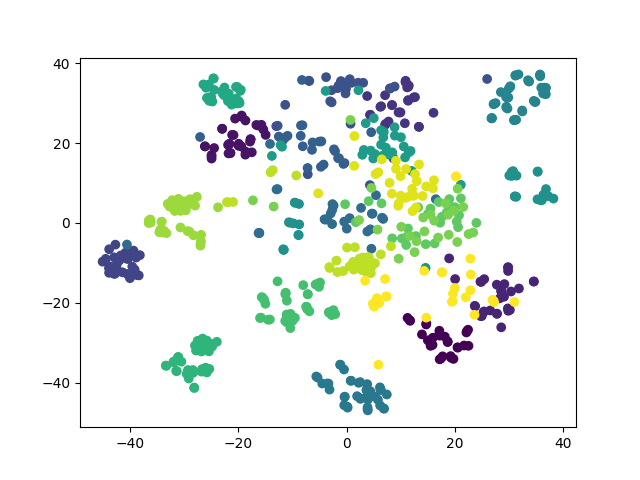}}
\subfigure[Adv. Contrastive]{\label{fig:tsne-d}\includegraphics[width=4.2cm,trim={0.5cm 0.5cm 0.5cm 0.5cm}, clip]{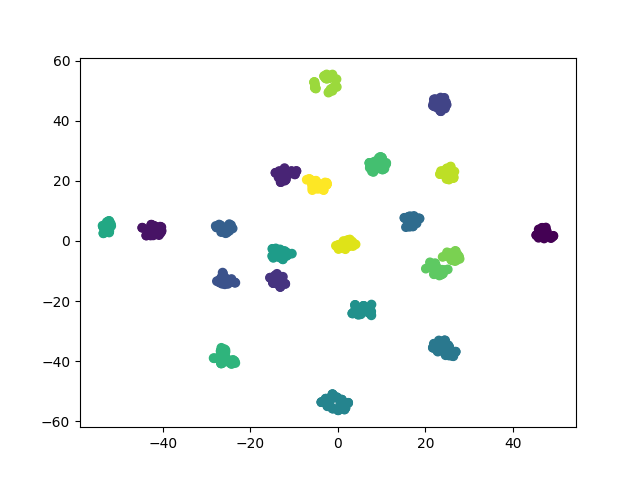}}
\caption{T-SNE plots on the JHMDB dataset using I3D features. Figure~\ref{fig:tsne-a} shows features (circles) and their respective adversarial negative samples as `x's (shown for 5 classes). Each color corresponds to a distinct class. Note that the negative samples (`x's) for each class are very close in embedding to the (positive) data samples (circles). In Figure~\ref{fig:tsne-b}, we plot the embeddings of sequence representations produced via average pooling of frame-level features (for all 21 classes). In Figure~\ref{fig:tsne-c}, we use our contrastive optimal transport for representation learning, however uses random noise to contrast against. In Figure~\ref{fig:tsne-d}, we embed the representations produced via adversarially-constrastive OT (ACOT). Our scheme results in well-separated clusters, justifying the superior performances.}
\label{fig:tsne}
\end{figure*}

\subsection{Ablative Studies}
In Table~\ref{tab:ablative}, we provide a thorough ablative study of the various modules in our framework. Specifically, we compare (i) average pooling, which provides the baseline performance on the dataset, (ii) Contrastive Optimal Transport (COT) without the adversarial noise, instead using Gaussian noise (iii) adversarially contrastive optimal transport (ACOT), (iv) ACOT with the distortion penalty (PCA) as characterized via a data reconstruction loss, (v) adversarially contrastive estimation with the PCA and temporal ordering constraints, but without the optimal transport cost, and (vi) our full model (ACOT+PCA+temporal order). For (ii), we found that using random noise that is completely uncorrelated with the data resulted in very poor performance (COT+Random). For this experiment, we produced the negative samples by multiplying random noise with the maximum feature value and subtracting it from the feature. That is, for a feature $\vx$, we generate $\vz\sim \normal(0, \max(\vx))$, and compute the negative examples $\vy=\gamma(\vx-\vz)$. Here, $\max(\vx)$ computes the maximum value in the dimensions of $\vx$. We repeated this ablative study on the two datasets over vgg and I3D features.

As is clear from the Table~\ref{tab:ablative}, using COT leads to better performances than average pooling in most cases, with ACOT demonstrating better performance than COT+Random in most cases. We found that there is a significant variance ($\pm$2\%) on COT+Random and thus our numbers are averaged over 5 trials. For ACOT, we sampled twice the number of negative samples as the positives. The combinations of PCA and order also demonstrates benefits overall, showing the effectiveness of our proposed method in extracting weak temporal signals from the sequential data. On the JHMDB dataset, our full architecture is significantly better than average pooling by nearly 7\% on both vgg and I3D features. A similar observation is made on the HMDB dataset as well. Our full model is about 2-5\% better on the RGB and FLOW streams separately and about 2\% better in combination. This clearly demonstrates the generalizability of our method to different datasets, types of features, and data modalities (RGB and optical flow).

\noindent\textbf{Increasing Number of Subspaces:}
In Figure~\ref{fig:subspaces}, we keep the distortion $\beta_1=0.1$ and number of negatives to 50, and plot the performance of ACOT against increasing number of subspaces in $\mU$. As we see, there is an increasing trend in the performance (on the HMDB dataset). However, as the number of subspaces increases, the representation learning time also increases (about $k$ times slower for $k$ subspaces). Beyond 8 subspaces, we did not find any improvements.

\noindent\textbf{Increasing Distortion Penalty:}
Keeping $\#$-subspaces at 1, and $\#$-negatives at 50, we increased $\beta_1$ from 0 to 2. As seen in Figure~\ref{fig:pca_val}, we see a marginal improvement in performance with increasing $\beta_1$. We believe, the contrastive formulation is already capturing useful properties of the input sequences that the contribution from an explicit distortion penalty is incremental.

\noindent\textbf{Increasing Number of Negatives:}
An advantage of our setup is the possibility of generating unlimited number of negatives.  In Figure~\ref{fig:num_negs}, we increased the number of negative samples from 20 to 200. While, it is very expensive for the OT to solve for large negative sets, we found that using about 40-100 samples is adequate and demonstrates performance improvements. However, with a large number of negatives (such as 200), counter-intuitively, we find that the accuracy drops significantly. This is perhaps because our noise generator model does not fool the classifier perfectly; as a result, overabundant negatives may be overlapping in distribution to the positives; diminishing the contrastiveness. 

\noindent\textbf{Increasing Temporal Ranking Regularization:}
In this experiment, we kept $\beta_1=0.1$, number of negatives to 50, and number of subspaces to 1, and changed the temporal regularization $\beta_2$ (in~\eqref{eq:full-objective}). Figure~\ref{fig:beta2} plots this sensitivity. As is clear from the figure, smaller regularization is ineffective.

\textbf{Running Time:} On average, excluding the time to extract CNN features, our representation learning setup takes about 30 frames per second using 5 iterations of conjugate gradient and 1000 iterations of inexact proximal optimal transport.

\textbf{Qualitative Visualizations:}
To gain insights into the kind of perturbations our adversarial network generates, we trained this sub-module on the CIFAR10 dataset. Specifically, we use an auto-encoder on the CIFAR images, the latent vectors of each image forms the feature. Next, we trained our adversarial network, similar to our setup for sequences, but assuming only a single frame (the encoded CIFAR image). We combined the generated noise with the latent feature and decoded the image. The goal of the discriminator in our framework is to classify this generated image as ``fake'', while the generator is trained to produce better noise such that the discriminator is fooled, however, a pre-trained CIFAR classifier (trained on the 10 CIFAR classes) shows a low-confidence against the true image class (i.e., the noise needs to be adversarial). In Figure~\ref{fig:cifar}, we show a few qualitative CIFAR images. The figure shows the noise impacts image regions that are likely to be useful for recognition (e.g., the top-right cat image, and its respective perturbation that corrupts mainly the face region). Figure~\ref{fig:tsne} shows T-SNE embeddings of representations produced by our scheme on JHMDB features. 

\subsection{Comparisons to the State of the Art}
In Tables~\ref{tab:soa_jhmdb} and~\ref{tab:soa_comparisons}, we compare the performances of our method against prior works on the respective datasets, such as (i) generalized rank pooling (GRP)~\cite{cherian2017generalized}, (ii) kernelized rank pooling (KRP)~\cite{cherian2018non}, and P-CNN~\cite{cheron2015p}. Our method performs better by about 2\% on three-fold cross-validation. We repeated this experiment using I3D features and compare against similar prior methods. Our results demonstrating superior performance.  For experiments on the HMDB dataset, we fine-tuned the I3D model on this dataset (this is in contrast to the results reported in Table~\ref{tab:ablative} that used a pre-trained I3D model). For our ACOT, we used a single hyperplane subspace. Our re-trained model produces a baseline 3-split accuracy of 80.6\% (we report the numbers from~\cite{carreira2017quo} in Table~\ref{tab:soa_comparisons}), and our proposed approach improves it to 81.8\%. We also compare to the results in~\cite{Wang_Cherian2018} that uses a similar pooling setup via adversarial perturbations for generating the negative set; our proposed model shows benefits.

\begin{table}[]
\begin{center}
\begin{tabular}{l|c}
    \hline
    JHMDB using VGG features & Accuracy\\
    \hline
    GRP~\cite{cherian2017generalized} & 70.6 \\
    P-CNN~\cite{cheron2015p} &  72.2\\
    Kernelized Pooling~\cite{cherian2018non} & 73.8 \\
    Ours (full model) & \textbf{75.7}\\
    \hline
    JHMDB using 3D-CNNs & Accuracy\\
    \hline
    Chained~\cite{Zolfaghari_2017_ICCV} & 76.1 \\
    I3D + Potion~\cite{choutas2018potion} & 85.5 \\
    I3D + Ours (full model) & \textbf{87.5}\\
    \hline
\end{tabular}
\vspace*{-0.5cm}
\end{center}
\caption{Comparisons on JHMDB dataset (3-splits).}
\label{tab:soa_jhmdb}
\vspace*{-0.3cm}
\end{table}
\begin{table}[]
    \centering
    \begin{tabular}{c|c}
        Method & Acc. (\%)\\
        \hline
       I3D ~\cite{carreira2017quo}  & 80.9  \\
       Disc. Pool~\cite{Wang_Cherian_TPAMI2019} & 81.3 \\
       DSP~\cite{Wang_Cherian2018} & 81.5\\
       \hline
         Ours (I3D+full model) & \textbf{81.8} \\
    \hline
    \end{tabular}
    \vspace*{-0.3cm}
    \caption{Comparisons on the HMDB dataset (3 splits) using two-stream I3D model fine-tuned on the dataset.}
    \label{tab:soa_comparisons}
    \vspace*{-0.3cm}
\end{table}

\section{Conclusions}
\label{sec:conclude}
We presented a novel framework for producing data representations on sequences via contrastive learning. Our key insight to look at this problem emerged from the observation that each item in a (video) sequence is often encoded using a model that does not access the full sequence. As a result, the cues for sequence level inference within such encodings might be weak. To amplify such cues, we resorted to contrastive learning, where we contrast the features against adversarially-learned features, and learns subspaces, as a representation, that captures these weak cues via optimal transport. We presented experiments on two datasets, demonstrating empirical benefits against recent methods.  

\section*{Acknowledgements}
Anoop Cherian thanks Matthew Brand and Anshul Shah for helpful suggestions. Shuchin Aeron acknowledges the support by NSF CCF:1553075 and support by MERL/Tufts University during the sabbatical period Jan 2019-August 2019. 
\appendix
%\section{Supplementary Material}
\section{Proof of Theorem 1}
\label{sec:proof}
\begin{proof}
Here we will prove a slightly general form of Theorem 1. We begin by noting that, 
\begin{align*}
    \mc{P}^2_k &\triangleq  \max_{\M{U}: \mc{S}(d,k)}    \min_{\pi\in \Pi(\mu_{\mX}, \nu_{\mY})} \mb{E}_{\pi} \| \M{U}^\top \V{x} - \M{U}^\top \V{y}\|^2,\\
    & = \max_{\M{U}: \mc{G}(d,k)}    \min_{\pi\in \Pi(\mu_{\mX}, \nu_{\mY)})} \mb{E}_{\pi} \| \M{U}\M{U}^\top \V{x} - \M{U} \M{U}^\top \V{y}\|^2 \\\text{ and } \\\mc{S}_k^2 & \triangleq \min_{\pi\in\Pi(\mu_{\mX}, \nu_{\mY})} \max_{\mU\in\mc{S}(d,k)} \mb{E}_{\pi} \| \M{U}^\top \V{x} - \M{U}^\top \V{y}\|^2.\\
    & = \min_{\pi\in\Pi(\mu_{\mX}, \nu_{\mY})} \max_{\mU\in\mc{G}(d,k)} \mb{E}_{\pi} \| \M{U} \M{U}^\top \V{x} - \M{U} \M{U}^\top \V{y}\|^2.
\end{align*}
 Now, 
\begin{align}
      &  \mc{C}_k^2 = \max_{\mU\in\grassmann(d,k)} \min_{\pi\in\Pi(\mu_{\mX}, \nu_{\mY})} \mb{E}_{\pi} \|\mU\mU^{\top} \V{x} - \V{y}\|^2 \notag \\
       & = \max_{\mU\in\grassmann(d,k)} \min_{\pi\in\Pi(\mu_{\mX}, \nu_{\mY})} \mb{E}_{\pi} \{ \|\mU\mU^{\top} \V{x} - \mU\mU^{\top} \V{y} \|^2+ \nonumber \\ & \hspace{45mm} \|(\M{I}_d - \mU\mU^{\top}) \V{y} \|^2\}\notag\\
       & \geq \max_{\mU\in\grassmann(d,k)} \min_{\pi\in\Pi(\mu_{\mX}, \nu_{\mY})} \mb{E}_{\pi} \|\mU\mU^{\top} \V{x} - \mU\mU^{\top} \V{y}\|^2 \\
       & =  \mc{P}_k^2 
\end{align}
Now since $\max \min \leq \min \max$,
\begin{align}
    & \max_{\mU\in\grassmann(d,k)} \min_{\pi\in\Pi(\mu_{\mX}, \nu_{\mY})} \mb{E}_{\pi}  \|\mU\mU^{\top} \V{x} - \V{y}\|^2 \notag  \\ 
    & \leq  \min_{\pi\in\Pi(\mu_{\mX}, \nu_{\mY})} \max_{\mU\in\grassmann(d,k)} \{ \mb{E}_{\pi}  \|\mU\mU^{\top} \V{x}_i - \mU\mU^{\top} \V{y}_j\|^2 + \nonumber \\ & \hspace{35mm} \mb{E}_{\pi} \|(\M{I}_d - \mU\mU^{\top}) \V{y} \|^2 \}\notag\\
    &\leq \mc{S}_k^2 + \max_{\mU\in\grassmann(d,k)} \mb{E}_{\nu_{\M{Y}}} \|(\M{I}_d - \mU\mU^{\top}) \V{y}\|^2.
\end{align}
The term $\max_{\mU\in\grassmann(d,k)} \mb{E}_{\nu_{\M{Y}}} \|(\M{I}_d - \mU\mU^{\top}) \V{y}\|^2 = \sum_{\ell = k+1}^{d} e_\ell(\M{\Sigma}_{\M{Y}})$ where $e_1, e_2, ..., e_d$ are the eigenvalues of the Gram matrix arranged in increasing order.
\end{proof}

\begin{figure*}[ht]
    \centering
    \subfigure[RGB]{\label{fig:fooling-a}\includegraphics[width=5.5cm,trim=0.2cm 6.2cm 0.2cm 5cm,clip]{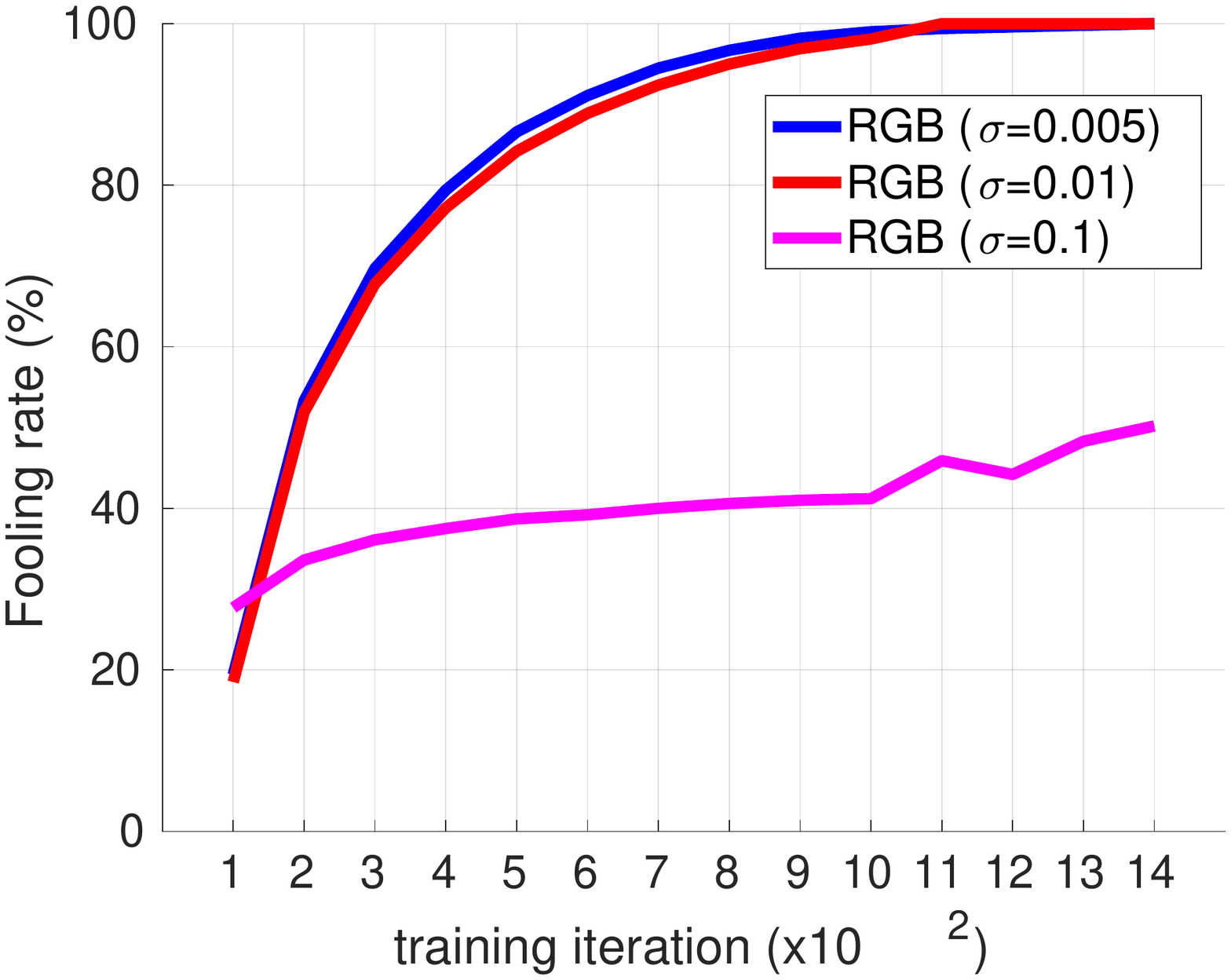}}
    \subfigure[FLOW]{\label{fig:fooling-b}\includegraphics[width=5.5cm,trim=0.2cm 6.2cm 0.2cm 5cm,clip]{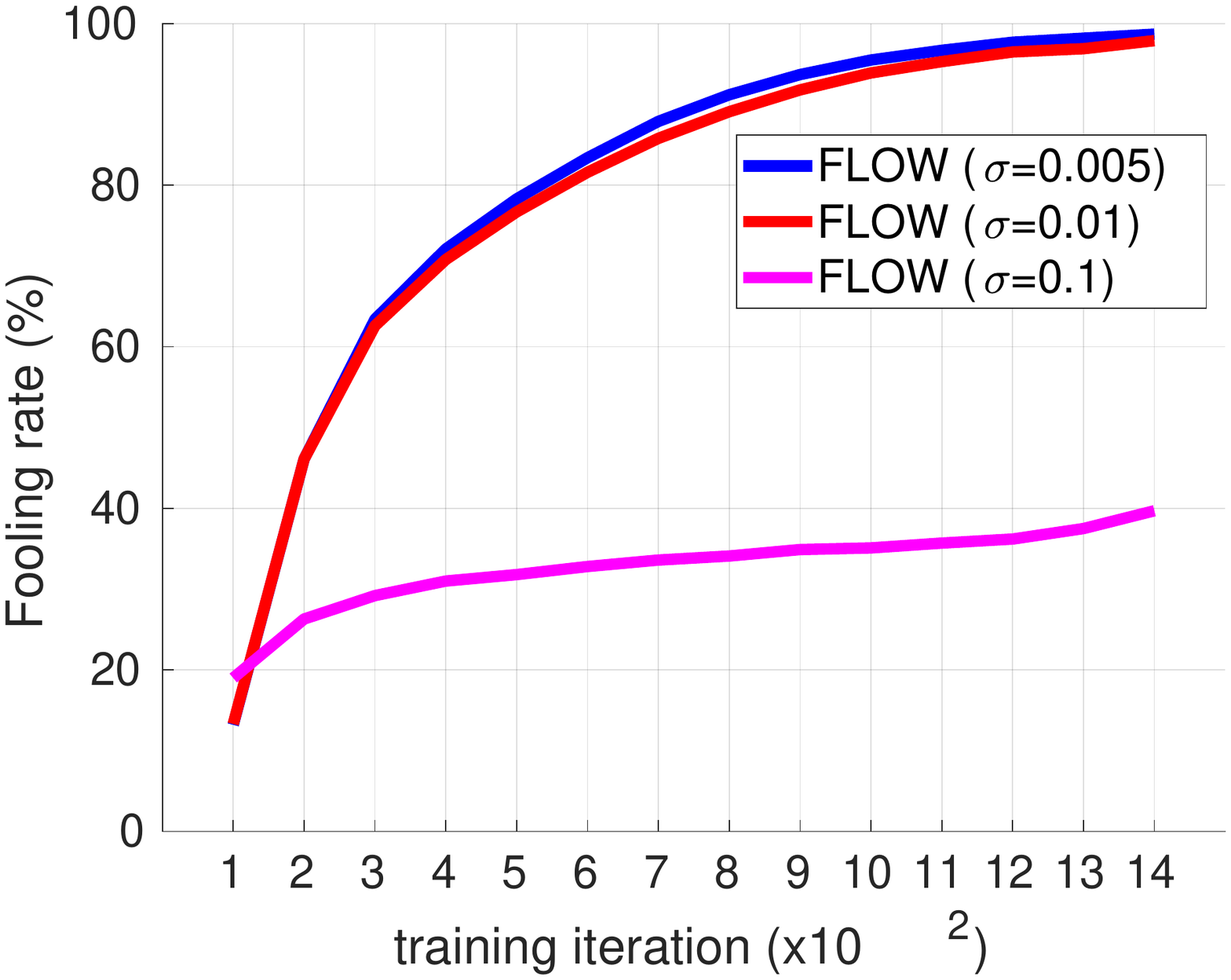}}
    \subfigure[ACOT Accuracy]{\label{fig:fooling-acc}\includegraphics[width=5.5cm,trim=0.2cm 6.2cm 0.2cm 5cm,clip]{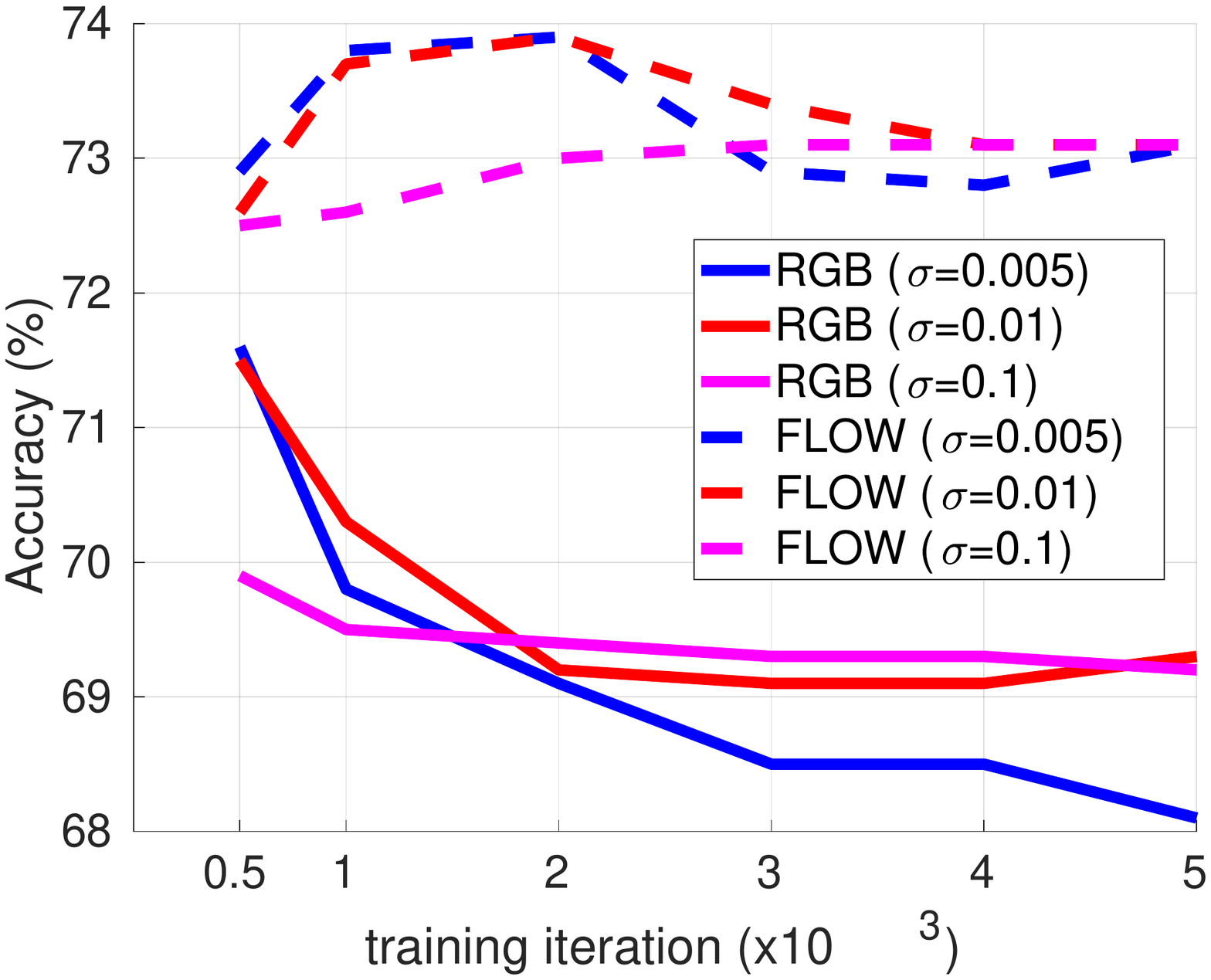}}
    \caption{Fooling rates for RGB stream (Figure~\ref{fig:fooling-a}) and FLOW stream (Figure~\ref{fig:fooling-b}) using I3D network on the HMDB dataset against the number of WGAN training iterations. We plot for three different variances of the Normal distribution, i.e., $\sigma=0.005, 0.01, 0.1$. Note that standard deviation of the features is about 0.008. As we see from the two plots, with a lower $\sigma=0.005,0.01$, the WGAN learns to generate adverarial pertubations with 100\% fooling rate in about 1000 iterations, however with a larger $\sigma=0.1$, the network could achieve about 50\% fooling rate on average. On the right~\ref{fig:fooling-acc}, we plot the validation accuracy (of ACOT) against the respective training iterations on the left. For RGB, higher-fooling rates seem to affect performances, however, the effect is reversed on the FLOW stream. This is perhaps because the RGB stream of I3D does not capture any useful temporal cues.}
    \label{fig:fooling-rates}
\end{figure*}

\section{Additional Experiments}
In this section, we detail our neural architectures in our COT framework and provide ablative studies of the various choices in our setup. 

\noindent\textbf{Datasets and Features:} As noted in the main paper, we use two datasets, namely (i) the JHMDB dataset, and (ii) the HMDB dataset. For the former, we explore our scheme using two types of features: (i) vgg-16 features, and (ii) I3D features. The vgg-16 features are 4096 dimensional each for every frame in the sequence. That is, we have feature matrices of size $4096\times n$ and $4096\times n-1$ for the RGB and optical flow respectively, where $n$ denotes the number of frames in the sequence. As for the I3D features, they are 1024 dimensional each and are extracted from the average pooling layer (after the ``$max\_5c$'' layer) of the Inception V3 network~\cite{carreira2017quo}. These features are produced from short clips, in which the I3D network takes clips consisting of 8 consecutive video frames, and produces one 1024 dimensional feature for that short clip. We use a sliding window with a temporal stride of 2 frames to generate our feature matrix for the two streams. Thus, in our setup, for a sequence with $n$ frames, we will have feature matrices of size $1024\times \lfloor\frac{n}{2}\rfloor$ and $1024\times \lfloor \frac{n-1}{2}\rfloor$ for the RGB and flow streams respectively.  Note that the features (from either network) are the outputs of ReLU activations and thus are all non-negative. We also normalize these features to have unit-norm.

\noindent\textbf{Baseline Networks and Training:}
As alluded to in the main paper, we have not trained the baseline networks ourselves as our goal is to demonstrate the advantages of adversarially constrastive optimal transport on features extracted from off-the-shelf neural models. To this end, for the vgg-16 features on the JHMDB dataset, we directly use the features provided to us by the authors of~\cite{cherian2017generalized}. As is mentioned in that paper, these features were infact produced using a network that was fine-tuned on the JHMDB dataset. For the I3D features, we used a ImageNet+Kinetics pre-trained I3D network implemented in PyTorch from a public git-hub repo\footnote{\url{https://github.com/piergiaj/pytorch-i3d}} to extract the features as described above. 

\subsection{Neural architectures}
Apart from the baseline feature-generating neural networks as described above, our framework has two other neural sub-modules, namely (i) the Wasserstein GAN (WGAN) framework for generating the adversarial samples, and (ii) the classifier to ensure the samples are adversarial. 

\noindent\textbf{Generator and Discriminator:}
Our generator $g$ has the following neural composition: 
\[g:= \left[\FCN(d, d), \ReLU(), \FCN(d, d), \ReLU(), \FCN(d, d)\right]\]
where $d$ is the input feature dimensionality (4096 for vgg-16 and 1024 for I3D), where this input is a noise sample from a multivariate normal distribution. Our discriminator has a similar structure, except that the final layer uses $\FCN(d,1)$. 

\noindent\textbf{Classifier:} As our representations for the sequences are linear subspaces, we decided to have the adversarial classifier also be limited in capacity, and thus we used a linear classifier for $\zeta$ in (10). Specifically, our classifier consists of a single $\FCN(d,c)$, where $c$ denotes the number of data classes. We attempted adding more layers and non-linearities to this classifier, however we found that such attempts made it difficult for the generator to learn the perturbations, and also the learned perturbations were difficult to be separated using the linear subspaces $U$ in our ACOT scheme.

\subsection{Adversarial Training}
We used RMSprop for training our models. We used a learning rate of $1e-4$ for the generator and discriminator, and for the classifier. We trained the classifier for 500 iterations and it achieves roughly 80\% accuracy on the input features (on the training set). More training resulted in overfitting, and thus posed difficulties when training the subsequent adversarial network. For WGAN, we adapted the public implementation from the authors of~\cite{arjovsky2017wasserstein}. This code uses 5 discriminative updates for every generator updates, which we also found to be useful in our setup. We measured the quality of the generated perturbations via the fooling rates on the positive samples. Specifically, the generated random perturbations are added to the original data samples (positives), passed through a $\ReLU()$, and then normalized to unit norm (note that all our data is unit-normalized) to produce the negative samples. Thus, if $c$ is the correct class label that a classifier $\zeta$ produces on an input $\vx$, then $\vy=\frac{\ReLU(\vx+g(\vz))}{\enorm{\ReLU(\vx+g(\vz))}}$, where $\vz\sim\normal(\bmX, \sigma^2\eye)$ is classified as $\bar{c}$ by $\zeta$, where $\bar{c}$ means the class $c$ has the lowest likelihood of being predicted, i.e., $c=\softmin(\zeta(\vy))$. We define fooling rate as the performance of the generator to produce a $\vy$ that fools $\zeta$ as described. Figure~\ref{fig:fooling-rates} show the trend in training the WGAN for various choices of $\sigma$ and its impact on the ACOT performance. Please see the text accompanying Figure~\ref{fig:fooling-rates} for the empirical analysis. Going by that analysis, we use $\sigma=0.01$ in our experiments.

\bibliography{acot}

\begin{thebibliography}{47}
\providecommand{\natexlab}[1]{#1}
\providecommand{\url}[1]{\texttt{#1}}
\expandafter\ifx\csname urlstyle\endcsname\relax
  \providecommand{\doi}[1]{doi: #1}\else
  \providecommand{\doi}{doi: \begingroup \urlstyle{rm}\Url}\fi

\bibitem[Absil et~al.(2009)Absil, Mahony, and Sepulchre]{absil2009optimization}
Absil, P.-A., Mahony, R., and Sepulchre, R.
\newblock \emph{Optimization algorithms on matrix manifolds}.
\newblock Princeton University Press, 2009.

\bibitem[Arjovsky et~al.(2017)Arjovsky, Chintala, and
  Bottou]{arjovsky2017wasserstein}
Arjovsky, M., Chintala, S., and Bottou, L.
\newblock Wasserstein gan.
\newblock \emph{arXiv preprint arXiv:1701.07875}, 2017.

\bibitem[Arora et~al.(2019)Arora, Khandeparkar, Khodak, Plevrakis, and
  Saunshi]{arora2019theoretical}
Arora, S., Khandeparkar, H., Khodak, M., Plevrakis, O., and Saunshi, N.
\newblock A theoretical analysis of contrastive unsupervised representation
  learning.
\newblock \emph{arXiv preprint arXiv:1902.09229}, 2019.

\bibitem[Bose et~al.(2018)Bose, Ling, and Cao]{Avishek_ACE}
Bose, A.~J., Ling, H., and Cao, Y.
\newblock Adversarial contrastive estimation.
\newblock \emph{arXiv preprint arXiv:1805.03642}, 2018.

\bibitem[Carreira \& Zisserman(2017)Carreira and Zisserman]{carreira2017quo}
Carreira, J. and Zisserman, A.
\newblock Quo vadis, action recognition? a new model and the kinetics dataset.
\newblock In \emph{CVPR}, 2017.

\bibitem[Cherian et~al.(2017)Cherian, Fernando, Harandi, and
  Gould]{cherian2017generalized}
Cherian, A., Fernando, B., Harandi, M., and Gould, S.
\newblock Generalized rank pooling for activity recognition.
\newblock In \emph{CVPR}, 2017.

\bibitem[Cherian et~al.(2018)Cherian, Sra, Gould, and Hartley]{cherian2018non}
Cherian, A., Sra, S., Gould, S., and Hartley, R.
\newblock Non-linear temporal subspace representations for activity
  recognition.
\newblock In \emph{CVPR}, 2018.

\bibitem[Ch{\'e}ron et~al.(2015)Ch{\'e}ron, Laptev, and Schmid]{cheron2015p}
Ch{\'e}ron, G., Laptev, I., and Schmid, C.
\newblock {P-CNN}: Pose-based {CNN} features for action recognition.
\newblock In \emph{ICCV}, 2015.

\bibitem[Choutas et~al.(2018)Choutas, Weinzaepfel, Revaud, and
  Schmid]{choutas2018potion}
Choutas, V., Weinzaepfel, P., Revaud, J., and Schmid, C.
\newblock Potion: Pose motion representation for action recognition.
\newblock In \emph{CVPR}, 2018.

\bibitem[Chung et~al.(2015)Chung, Kastner, Dinh, Goel, Courville, and
  Bengio]{RNN1}
Chung, J., Kastner, K., Dinh, L., Goel, K., Courville, A., and Bengio, Y.
\newblock A recurrent latent variable model for sequential data.
\newblock In \emph{NIPS}, 2015.

\bibitem[Cover \& Thomas(2006)Cover and Thomas]{cover2006}
Cover, T. and Thomas, J.
\newblock \emph{{Elements of Information Theory}}.
\newblock John Wiley and Sons, 2006.

\bibitem[Cuturi(2013)]{Cuturi_2013}
Cuturi, M.
\newblock Sinkhorn distances: Lightspeed computation of optimal transport.
\newblock In \emph{NIPS}, 2013.

\bibitem[{Dollar} et~al.(2005){Dollar}, {Rabaud}, {Cottrell}, and
  {Belongie}]{Dollar}
{Dollar}, P., {Rabaud}, V., {Cottrell}, G., and {Belongie}, S.
\newblock Behavior recognition via sparse spatio-temporal features.
\newblock In \emph{Intl. Workshop on Visual Surveillance and Performance
  Evaluation of Tracking and Surveillance}, 2005.

\bibitem[Fletcher \& Reeves(1964)Fletcher and Reeves]{fletcher1964function}
Fletcher, R. and Reeves, C.~M.
\newblock Function minimization by conjugate gradients.
\newblock \emph{The computer journal}, 7\penalty0 (2):\penalty0 149--154, 1964.

\bibitem[Girdhar et~al.(2019)Girdhar, Tran, Torresani, and
  Ramanan]{Girdhar_ICCV19}
Girdhar, R., Tran, D., Torresani, L., and Ramanan, D.
\newblock Distinit: Learning video representations without a single labeled
  video.
\newblock \emph{CoRR}, abs/1901.09244, 2019.
\newblock URL \url{http://arxiv.org/abs/1901.09244}.

\bibitem[Greff et~al.(2016)Greff, Srivastava, Koutn{\'\i}k, Steunebrink, and
  Schmidhuber]{greff2016lstm}
Greff, K., Srivastava, R.~K., Koutn{\'\i}k, J., Steunebrink, B.~R., and
  Schmidhuber, J.
\newblock Lstm: A search space odyssey.
\newblock \emph{IEEE Transactions on Neural Networks and Learning Systems},
  28\penalty0 (10):\penalty0 2222--2232, 2016.

\bibitem[Gutmann \& Hyv{\"a}rinen(2010)Gutmann and
  Hyv{\"a}rinen]{gutmann2010noise}
Gutmann, M. and Hyv{\"a}rinen, A.
\newblock Noise-contrastive estimation: A new estimation principle for
  unnormalized statistical models.
\newblock In \emph{AISTATS}, 2010.

\bibitem[Harandi et~al.(2014)Harandi, Salzmann, Jayasumana, Hartley, and
  Li]{harandi2014expanding}
Harandi, M.~T., Salzmann, M., Jayasumana, S., Hartley, R., and Li, H.
\newblock Expanding the family of grassmannian kernels: An embedding
  perspective.
\newblock In \emph{ECCV}, 2014.

\bibitem[H{\'{e}}naff et~al.(2019)H{\'{e}}naff, Razavi, Doersch, Eslami, and
  van~den Oord]{Henaff2019}
H{\'{e}}naff, O.~J., Razavi, A., Doersch, C., Eslami, S. M.~A., and van~den
  Oord, A.
\newblock Data-efficient image recognition with contrastive predictive coding.
\newblock \emph{CoRR}, abs/1905.09272, 2019.
\newblock URL \url{http://arxiv.org/abs/1905.09272}.

\bibitem[Hoffer \& Ailon(2015)Hoffer and Ailon]{hoffer2015deep}
Hoffer, E. and Ailon, N.
\newblock Deep metric learning using triplet network.
\newblock In \emph{International Workshop on Similarity-Based Pattern
  Recognition}, 2015.

\bibitem[Jhuang et~al.(2013)Jhuang, Gall, Zuffi, Schmid, and
  Black]{jhuang2013towards}
Jhuang, H., Gall, J., Zuffi, S., Schmid, C., and Black, M.~J.
\newblock Towards understanding action recognition.
\newblock In \emph{ICCV}, 2013.

\bibitem[Kim et~al.(2019)Kim, Ahn, and Bengio]{NIPS2019_9332}
Kim, T., Ahn, S., and Bengio, Y.
\newblock Variational temporal abstraction.
\newblock In \emph{NeurIPS}. 2019.

\bibitem[Kuehne et~al.(2011)Kuehne, Jhuang, Garrote, Poggio, and
  Serre]{kuehne2011hmdb}
Kuehne, H., Jhuang, H., Garrote, E., Poggio, T., and Serre, T.
\newblock {HMDB}: a large video database for human motion recognition.
\newblock In \emph{ICCV}, 2011.

\bibitem[Merity et~al.(2018)Merity, Keskar, and Socher]{merity2017regularizing}
Merity, S., Keskar, N.~S., and Socher, R.
\newblock Regularizing and optimizing lstm language models.
\newblock In \emph{ICLR}, 2018.

\bibitem[Moosavi-Dezfooli et~al.(2016)Moosavi-Dezfooli, Fawzi, and
  Frossard]{moosavi2016deepfool}
Moosavi-Dezfooli, S.-M., Fawzi, A., and Frossard, P.
\newblock Deepfool: a simple and accurate method to fool deep neural networks.
\newblock In \emph{CVPR}, 2016.

\bibitem[Oord et~al.(2018)Oord, Li, and Vinyals]{oord2018representation}
Oord, A. v.~d., Li, Y., and Vinyals, O.
\newblock Representation learning with contrastive predictive coding.
\newblock \emph{arXiv preprint arXiv:1807.03748}, 2018.

\bibitem[Paty \& Cuturi(2019)Paty and Cuturi]{Paty_Cuturi2019}
Paty, F. and Cuturi, M.
\newblock Subspace robust wasserstein distances.
\newblock \emph{CoRR}, abs/1901.08949, 2019.
\newblock URL \url{http://arxiv.org/abs/1901.08949}.

\bibitem[Poole et~al.(2019)Poole, Ozair, Van Den~Oord, Alemi, and
  Tucker]{pmlr-v97-poole19a}
Poole, B., Ozair, S., Van Den~Oord, A., Alemi, A., and Tucker, G.
\newblock On variational bounds of mutual information.
\newblock In \emph{ICML}, 2019.

\bibitem[Santambrogio(2015)]{OTAM_book}
Santambrogio, F.
\newblock Optimal transport for applied mathematicians.
\newblock \emph{Birk{\"a}user, NY}, 55\penalty0 (58-63):\penalty0 94, 2015.

\bibitem[Saunshi et~al.(2019)Saunshi, Plevrakis, Arora, Khodak, and
  Khandeparkar]{SaunshiPAKK19}
Saunshi, N., Plevrakis, O., Arora, S., Khodak, M., and Khandeparkar, H.
\newblock A theoretical analysis of contrastive unsupervised representation
  learning.
\newblock In \emph{ICML}, 2019.

\bibitem[Simonyan \& Zisserman(2014)Simonyan and Zisserman]{simonyan2014two}
Simonyan, K. and Zisserman, A.
\newblock Two-stream convolutional networks for action recognition in videos.
\newblock In \emph{NIPS}, 2014.

\bibitem[Smith \& Eisner(2005)Smith and Eisner]{smith2005contrastive}
Smith, N.~A. and Eisner, J.
\newblock Contrastive estimation: Training log-linear models on unlabeled data.
\newblock In \emph{ACL}, 2005.

\bibitem[Song \& Ermon(2019)Song and Ermon]{song2019understanding}
Song, J. and Ermon, S.
\newblock Understanding the limitations of variational mutual information
  estimators.
\newblock \emph{arXiv preprint arXiv:1910.06222}, 2019.

\bibitem[Sun et~al.(2019)Sun, Baradel, Murphy, and Schmid]{Sun2019}
Sun, C., Baradel, F., Murphy, K., and Schmid, C.
\newblock Contrastive bidirectional transformer for temporal representation
  learning.
\newblock \emph{CoRR}, abs/1906.05743, 2019.
\newblock URL \url{http://arxiv.org/abs/1906.05743}.

\bibitem[Sutskever et~al.(2014)Sutskever, Vinyals, and
  Le]{sutskever2014sequence}
Sutskever, I., Vinyals, O., and Le, Q.~V.
\newblock Sequence to sequence learning with neural networks.
\newblock In \emph{NIPS}, 2014.

\bibitem[Tran et~al.(2017)Tran, Ray, Shou, Chang, and Paluri]{Tran2017}
Tran, D., Ray, J., Shou, Z., Chang, S., and Paluri, M.
\newblock Convnet architecture search for spatiotemporal feature learning.
\newblock \emph{CoRR}, abs/1708.05038, 2017.
\newblock URL \url{http://arxiv.org/abs/1708.05038}.

\bibitem[Tschannen et~al.(2019)Tschannen, Djolonga, Rubenstein, Gelly, and
  Lucic]{Tschannen2019}
Tschannen, M., Djolonga, J., Rubenstein, P.~K., Gelly, S., and Lucic, M.
\newblock On mutual information maximization for representation learning.
\newblock \emph{arXiv preprint arXiv:1907.13625}, 2019.

\bibitem[van~den Oord et~al.(2018)van~den Oord, Li, and Vinyals]{Oord2018}
van~den Oord, A., Li, Y., and Vinyals, O.
\newblock Representation learning with contrastive predictive coding.
\newblock \emph{CoRR}, abs/1807.03748, 2018.

\bibitem[Varol et~al.(2016)Varol, Laptev, and Schmid]{VarolLS16}
Varol, G., Laptev, I., and Schmid, C.
\newblock Long-term temporal convolutions for action recognition.
\newblock \emph{CoRR}, abs/1604.04494, 2016.

\bibitem[Varol et~al.(2017)Varol, Laptev, and Schmid]{varol2017long}
Varol, G., Laptev, I., and Schmid, C.
\newblock Long-term temporal convolutions for action recognition.
\newblock \emph{IEEE Transactions on Pattern Analysis and Machine
  Intelligence}, 40\penalty0 (6):\penalty0 1510--1517, 2017.

\bibitem[Wang \& Cherian(2018)Wang and Cherian]{Wang_Cherian2018}
Wang, J. and Cherian, A.
\newblock Learning discriminative video representations using adversarial
  perturbations.
\newblock \emph{CoRR}, abs/1807.09380, 2018.
\newblock URL \url{http://arxiv.org/abs/1807.09380}.

\bibitem[{Wang} \& {Cherian}(2019){Wang} and {Cherian}]{Wang_Cherian_TPAMI2019}
{Wang}, J. and {Cherian}, A.
\newblock Discriminative video representation learning using support vector
  classifiers.
\newblock \emph{IEEE Transactions on Pattern Analysis and Machine
  Intelligence}, pp.\  1--1, 2019.
\newblock \doi{10.1109/TPAMI.2019.2937292}.

\bibitem[Wang \& Gupta(2015)Wang and Gupta]{WangG15a}
Wang, X. and Gupta, A.
\newblock Unsupervised learning of visual representations using videos.
\newblock \emph{CoRR}, abs/1505.00687, 2015.

\bibitem[Wu et~al.(2017)Wu, Zaheer, Hu, Manmatha, Smola, and
  Kr{\"{a}}henb{\"{u}}hl]{Wu2017}
Wu, C., Zaheer, M., Hu, H., Manmatha, R., Smola, A.~J., and
  Kr{\"{a}}henb{\"{u}}hl, P.
\newblock Compressed video action recognition.
\newblock \emph{CoRR}, abs/1712.00636, 2017.

\bibitem[Xie et~al.(2019)Xie, Wang, Wang, and Zha]{ipot}
Xie, Y., Wang, X., Wang, R., and Zha, H.
\newblock A fast proximal point method for computing exact wasserstein
  distance.
\newblock In \emph{UAI}, 2019.

\bibitem[Zilly et~al.(2017)Zilly, Srivastava, Koutn{\i}k, and
  Schmidhuber]{zilly2017recurrent}
Zilly, J.~G., Srivastava, R.~K., Koutn{\i}k, J., and Schmidhuber, J.
\newblock Recurrent highway networks.
\newblock In \emph{ICML}, 2017.

\bibitem[Zolfaghari et~al.(2017)Zolfaghari, Oliveira, Sedaghat, and
  Brox]{Zolfaghari_2017_ICCV}
Zolfaghari, M., Oliveira, G.~L., Sedaghat, N., and Brox, T.
\newblock Chained multi-stream networks exploiting pose, motion, and appearance
  for action classification and detection.
\newblock In \emph{ICCV}, 2017.

\end{thebibliography}
\bibliographystyle{icml2020}
% \newpage
% \hspace{5mm}
\end{document}